 \documentclass[english, ds]{agh-wi} 
\titlePL{}  
\titleEN{Application of predictive machine learning in pen\&paper RPG game design} 
\author{Jolanta Śliwa}
\supervisor{PhD Wojciech Czech}
\cosupervisor{Jakub Adamczyk}
\usepackage{polski} 
\usepackage{polski} 
\usepackage[sorting=none]{biblatex}           
\usepackage{amsmath}                          
\usepackage{amssymb}                          
\usepackage[polish, intoc]{nomencl}           
\usepackage{graphicx}                         
\PassOptionsToPackage{dvipsnames}{xcolor}
\usepackage{xcolor}                          
\usepackage{tabularx}                         
\usepackage{longtable}                        
\usepackage[ruled,linesnumbered]{algorithm2e} 
\usepackage{listings}                         
\usepackage[frozencache=true,cachedir=_minted-main]{minted}               
\usepackage{hyperref}                         
\usepackage{csquotes}                         
\usepackage{glossaries}
\usepackage{float}
\usepackage{dsfont}
\usepackage{booktabs}
\usepackage[margin=1in]{geometry}
\usepackage{svg}
\usepackage{colortbl}
\usepackage{multirow}
\usepackage{rotating}
\usepackage{makecell}
\usepackage{bm}
\usetikzlibrary{shapes.geometric,arrows.meta} 
\tikzstyle{every picture}+=[remember picture]
\tikzstyle{na} = [baseline=-.5ex]
\definecolor{shadecolor}{gray}{0.9}
\makenomenclature 

\newcounter{comment}[chapter]

\makeglossaries
\begin{document}
\frontmatter 
\maketitle 
\cleardoublepage
\thispagestyle{empty}
\vspace*{\fill}
\begin{flushright}
    \em
    \begin{minipage}{0.75\textwidth}
        
        I would like to thank my supervisors, Wojciech Czech and Jakub Adamczyk, for their valuable advice, guidance, and support throughout this work. I am especially grateful for their shared expertise in both machine learning methodologies and domain-specific knowledge, which significantly contributed to the development and direction of this project.
        
        \vspace{3\baselineskip}
        
    \end{minipage}
\end{flushright}
\vspace{3\baselineskip}

In the preparation of this master's thesis, AI-powered language tools were utilized to enhance the clarity, coherence, and overall quality of the written text. These tools assisted in tasks such as grammar correction, style refinement, and suggested phrasing. However, the content, ideas, research findings, and conclusions presented herein are solely the author's own. Any errors or omissions remain the responsibility of the author.
\begin{abstractPL}

    W ostatnich latach rynek gier fabularnych typu pen\&paper odnotował dynamiczny rozwój. W związku z tym firmy coraz częściej badają możliwości integracji technologii sztucznej inteligencji w celu zwiększenia satysfakcji graczy oraz uzyskania przewagi konkurencyjnej.

    Jednym z kluczowych wyzwań, przed którymi stoją wydawcy, jest projektowanie nowych przeciwników i szacowanie poziomu ich trudności. Obecnie nie istnieją konkurencyjne zautomatyzowane metody określania poziomu potworów; stosowane podejścia opierają się wyłącznie na testach manualnych i ocenie ekspertów. Choć metody te mogą dostarczać stosunkowo dokładnych oszacowań, są czasochłonne i wymagają dużych nakładów zasobów.
    
    Szacowanie poziomu trudności można zrealizować za pomocą technik regresji uporządkowanej (ang. \textit{ordinal regression}). Niniejsza praca przedstawia przegląd i ocenę metod stosowa-
    nych w tym zadaniu. Opisano również proces budowy dedykowanego zbioru danych służącego do estymacji poziomu. Ponadto opracowano model inspirowany ludzkim podejściem, który pełni rolę punktu odniesienia, umożliwiając porównanie algorytmów uczenia maszynowego z metodami tradycyjnie stosowanymi przez wydawców gier pen\&paper. Dodatkowo zaprojektowano specjalistyczną procedurę oceny opartą na wiedzy dziedzinowej, mającą na celu ocenę skuteczności modeli oraz umożliwienie trafnych porównań.
\end{abstractPL}
\begin{abstractEN}
    
    In recent years, the pen\&paper RPG market has experienced significant growth. As a result, companies are increasingly exploring the integration of AI technologies to enhance player experience and gain a competitive edge.

    One of the key challenges faced by publishers is designing new opponents and estimating their challenge level. Currently, there are no automated methods for determining a monster’s level; the only approaches used are based on manual testing and expert evaluation. Although these manual methods can provide reasonably accurate estimates, they are time-consuming and resource-intensive.
    
    Level prediction can be approached using ordinal regression techniques. This thesis presents an overview and evaluation of state-of-the-art methods for this task. It also details the construction of a dedicated dataset for level estimation. Furthermore, a human-inspired model was developed to serve as a benchmark, allowing comparison between machine learning algorithms and the approach typically employed by pen\&paper RPG publishers. In addition, a specialized evaluation procedure, grounded in domain knowledge, was designed to assess model performance and facilitate meaningful comparisons.
    
\end{abstractEN}
\tableofcontents   

\mainmatter 
\chapter{Introduction}

A role-playing game (RPG) is a type of game in which participants play as fictional characters that participate in a story narrated by Game Master (GM). This thesis focuses on pen \& paper RPGs, also known as tabletop RPGs (TTRPG).
In a TTRPG, players gather around a table, either physically or virtual, to create and develop characters who participate in a story that evolves over time \cite{ttrpg}. The game is based on the narration, guides and resolution of conflicts using rules systems and dice rolls. TTRPGs are highly dependent on imagination, collaboration, and improvisation.

Pen \& paper RPGs are distinct from computer RPGs (cRPGs) and live-action role-playing games (LARP), where participants dress up and physically perform the actions of their characters.  Here, the acronym RPG refers specifically to pen \& paper RPGs. 
In LARP, players not only create characters, but also physically act out their roles during gameplay \cite{larp}. They often create costumes to dress as their characters and use props. These games can range from small private RPG sessions to large-scale events with hundreds or even hundreds of participants.
In contrast, computer RPGs are a digital alternative to traditional RPGs \cite{crpg}. Unlike TTRPGs, which require extensive imagination and verbal storytelling, cRPGs present a structured game world in which players control characters through a graphical interface. These games follow pre-programmed narratives and mechanics, allowing for automated conflict resolution, exploration, and progression within a virtual environment.

TTRGP market has experienced significant growth in recent years. In 2023, the market was valued at approximately \$1.92 billion USD, with projections estimating that it will reach \$5.27 billion USD by 2033 \cite{ttrpg-growth}. This rapid expansion highlights the increasing popularity of TTRPGs and the evolving demands of players.
To remain competitive, publishers, miniature manufacturers, and other industry players must continuously innovate. Beyond creative marketing strategies and engaging gameplay demonstrations, companies place a strong emphasis on R\&D and technological advancements to improve the gaming experience. 
In addition, the TTRPG community is predominantly made up of younger players who are highly receptive to digital tools and technological solutions. Even when playing in person, many use on-line platforms for combat management, virtual dice rolling, etc.
Like any rapidly evolving market, the TTRPG industry is exploring the integration of artificial intelligence (AI) to gain an advantage. Companies are looking to AI-driven systems to improve player experiences.
One of the biggest players in the industry, Hasbro, the parent company of Wizards of the Coast—publisher of Dungeons \& Dragons (DnD), the most popular TTRPG series, has acknowledged plans to incorporate AI-based systems. The company has already begun using AI during the development of games such as DnD and Magic: The Gathering \cite{dnd-ai}. This shift shows the broader potential for ML-based tools to improve gameplay mechanics, enhance storytelling, and ensure balanced gameplay and opponent interactions.

In pen \& paper RPGs, gameplay revolves around the Game Master's (GM) narrative. The rest of the group plays the role of heroes with attributes such as charisma, strength, and intelligence, which change as the story progresses. Action takes place in a fantasy, fictional world inside the players' imagination. Game Master prepares adventures/scenarios and challenges for players and creates the main plot of the story. The main aim of players is to complete the scenario according to the set of rules, named the game system or mechanics. For example, their goals may include defeating a dragon, rescuing a princess, or kidnapping her.

One of the most important components of adventures is combat with the so-called monsters introduced by the GM. Monsters possess attributes similar to the players' characteristics which define their strength and difficulty. In many RPGs, the difficulty of the monster is indicated by a value called level. Selecting monsters of an appropriate level is essential to prepare encounters that are challenging but winnable to players. The difficulty of the encounter is calculated using publicly available formulas that depend on the levels of both players and monsters. Thus, determining a monster's level is crucial for encounter design. As monsters are described using numerical statistics, this is a natural task for ML-based solutions.

This thesis focuses on Pathfinder Second Edition, published by Paizo. As one of the most popular TTRPG systems, it has consistently ranked among the top three since its release. For about two years, Pathfinder held the number one position and most frequently occupies the second spot \cite{top5rpg}. Furthermore, it has received numerous awards, particularly from players \cite{readers-choice, tabletop-awards-2022}, and has fostered a strong and dedicated community.  Quoted a Paizo staff member: "Pathfinder 2e is the best-selling, most successful thing Paizo has ever made" \cite{reddit-tabletop-awards-2022}.

A key factor contributing to Pathfinder's success is its commitment to open-source accessibility. Monster datasets are available under the Open Game License 1.0A \cite{paizo-pathfinder-license}. More recently, Paizo has transitioned to publishing new scenarios and books under the Open RPG Creative (ORC) license \cite{pathfinder-orc}. This new licensing framework provides even stronger guarantees of accessibility and creative freedom \cite{pathfinder-orc-final-version}, further strengthening the appeal of the game and long-term sustainability within the TTRPG ecosystem.

\section{Research goals}

The aim of this work is to investigate techniques for monster level estimation to support the design of opponents for the pen-and-paper RPG Pathfinder Roleplaying Game Second Edition. The specific goals of the research project presented in this thesis are as follows:
\begin{enumerate}
\item implement an accurate and efficient automated estimation of monster level, based on its characteristics,
\item design realistic evaluation procedure, based on domain knowledge,
\item compare the efficiency of various ML algorithms for regression and ordinal regression.
\end{enumerate}

\section{Motivation}

Currently, there are no competitive methods to automatically determine the level of a monster. The only methods used are manual testing and expert assessment.

The manual method provides a relatively accurate level estimation, as testers engage with opponents in actual gameplay. The major drawback of this method is that estimating the level of an opponent requires many hours of work and the participation of multiple people. A typical game session lasts from 2 to 6 hours and involves 4 players and a game master. Several game sessions are conducted for each newly created monster.

Expert assessment involves fewer participants but can be inaccurate due to human error. This method is also time-consuming as it requires manual comparison of the monster with existing ones, of which there are thousands.

The Pathfinder system provides guidelines \cite{building-creatures} on how to design monsters and adjust them to create a creature with a chosen level that could be used to estimate the level of the monster. The presented solution focuses only on a subset of characteristics that are common to almost all monsters. As the authors admit, the solution they present is not perfect, working best for monsters relying on physical strength in combat. Monsters capable of casting spells or relying on non-melee abilities require different adjustments, which are not described.  This method is most effective for low-level monsters, as their characteristics are fewer and less diverse compared to the more complex high-level monsters.

Currently, there are many guides available on how to design monsters. They cover topics such as building worlds with personalized adversaries, focusing on the emotions a creature is meant to evoke, and more \cite{worldanvil-monsters-design} \cite{edmcrae-monsters-design}. However, these guides do not provide a formula for accurately determining a monster's level, which is crucial for gameplay. After designing an opponent using these guides, the game master still needs to spend time searching for similar monsters \cite{edmcrae-monsters-design}, and most often resorts to a manual method of comparing the created monster to the official ones. Currently, there is no competitive automated method for determining the level in the market. Pathfinder itself includes a system for creating monsters, along with certain methods to estimate their level \cite{building-creatures}. However, the entire process is time-consuming, imprecise, and requires extensive knowledge of existing monsters. The manual relies on approximate values for monsters within a given type and level.

\section{Contributions}

This thesis presents several contributions to the problem of ordinal regression, particularly in the context of estimating levels based on RPG data. The main contributions are as follows:
\begin{enumerate}
    \item A novel dataset was created for an ordinal regression task, based on monsters' data published for the Pathfinder Second Edition pen \& paper RPG.
    \item  A comprehensive evaluation procedure was designed for ordinal regression tasks with chronologically structured data. This includes the introduction of a chronological data split and an expanding window approach inspired by time series evaluation techniques. In addition, a set of metrics was selected to account for the ordinal nature of the labels, the temporal dependencies, and the class imbalance in the dataset.
    \item A human-inspired baseline model, derived from Pathfinder rulebooks, was developed and evaluated on the constructed dataset.
    \item State-of-the-art models for ordinal regression were implemented and tested to address the level estimation problem. In addition, their results were then analyzed and compared.
\end{enumerate}


\section{Glossary of RPG-related terms}

\noindent
\textbf{Game Master (GM)} - person who runs the game. Their tasks include primarily narrating the game and ensuring that players follow the rules \cite{games-most-wanted,low-tech-industry, playing-roles, creation-of-narrative-ttrpg}.
\textbf{Level} - an integer representing overall power of a character or object inside the game's world, e.g. a player character/hero, a monster, or an item.  In the case of monsters, these represent the danger they pose to the players \cite{level}. \\
\textbf{Monster} - creature that players must face during gameplay, controlled by the GM. Like players, they have their own sets of characteristics that define their strength and skills \cite{encoding-monsters, creation-of-narrative-ttrpg}.\\
\textbf{Player} - person who plays the role of a fictional character within a fantasy world, making decisions, rolling dice, and interacting with the game world as guided by a GM and set of rules.\\
\textbf{Player character} - a fictional character that is controlled by a player, not by the GM; the player makes decisions, takes actions, and role-plays this character within the game's narrative and rules. A character has a set of statistics that define its skills and strength.\\
\textbf{RPG session} - an uninterrupted period of time when a group of players and a game master gather around a table (in person or remotely) to play a selected RPG game. A session can be thought of as an episode of a longer story, the completion of an entire scenario \cite{creation-of-narrative-ttrpg}.\\

\chapter{Literature review}

\section{RPGs and game design}

The majority of research and literature on RPG game design, particularly concerning monsters, mostly emphasizes the emotional and narrative aspects that creatures should evoke during the gameplay. For example, significant attention is given to what monsters symbolize, how they contribute to the game world, and their role in the story \cite{worldanvil-monsters-design, edmcrae-monsters-design}.

On the other hand, studies intersecting computer science and RPG typically focus on digital role-playing video games rather than pen \& paper RPGs \cite{arulraj2010adaptive}. Such studies rarely address challenges specific to TTRPGs.

The field of Computer Science offers limited contributions related to pen \& paper RPGs. Existing research that engages with this area often also focuses on storytelling \cite{drachen2009towards} or the integration of technology to improve traditional gameplay experiences \cite{tychsen2006making}. There remains a significant research gap to address more mechanical or game-balance aspects of pen \& paper RPGs, such as estimating monster levels.

Using the examples of Canada and the United States, we can observe that the pen-and-paper RPG market continues to grow, especially in recent years. In 2019, their sales increased significantly by 23\%, followed by another increase of 31\% in 2020, exceeding \$100 million for the first time \cite{ttrpgs-market-grew-2020}. Since 2013, the market has grown approximately seven times \cite{ttrpgs-market-2013-2020}.

Since the debut of its first edition, the Pathfinder system has consistently ranked among the top five most popular pen-and-paper RPGs (except in the fall of 2018, when the announcement of its second edition temporarily disrupted the rankings). For about two years, Pathfinder held the number one position and most frequently occupies the second spot \cite{top5rpg}.

In 2019, Pathfinder won Tabletop RPG of the Year: Readers' Choice award \cite{readers-choice}. The second edition of the game earned the title of BEST ROLEPLAYING GAME 2022, awarded by Tabletop Gaming \cite{tabletop-awards-2022}.

\section{Ordinal regression}

Ordinal regression (also called ordinal classification) is a type of regression used to analyze relationships where the dependent variable is an ordinal variable \cite{winship1984regression, burkner2019ordinal}. Ordinal variables represent categorical data with an inherent order, but the intervals between the categories are not necessarily equal or well defined. This characteristic makes ordinal regression distinct from other forms of regression or classification. It accounts for the ordered nature of the dependent variable without assuming a continuous distribution or equal spacing between levels.

The simplest approach to this problem is a regular regression, but with rounding the predictions to defined levels. Alternatively, dedicated models for ordinal regression can be used.
Each approach will be described below.

\subsection{Regression with rounding}

Classical regression methods, including techniques such as linear regression and Random Forest, are among the widely used approaches for addressing ordinal regression problems \cite{kramer2001prediction, shin2022moving}. Due to the characteristics of these methods, which are designed to predict continuous outcomes, a post-processing step is required to adapt their outputs to ordinal data. This step involves mapping the predicted continuous values to discrete ordinal categories, typically through a rounding function. 

In the work by Kramer et al. \cite{kramer2001prediction}, three different rounding strategies were proposed for this purpose: median, mode, and rounded mean. Among these, only the rounded mean is applicable to the problem addressed in this study. All strategies use outputs from multiple models, applying the median, mean, or mode to all returned outputs. Due to the nature of these methods, only the mean strategy considers the case where the final model output is not an integer and applies mathematical rounding to the results. Similarly, the study \cite{shin2022moving} adopts a rounding-to-the-nearest-neighbor strategy, which is equivalent to classical mathematical rounding in this case.

\subsection{Dedicated models}

Ordinal regression gave rise to multiple models specifically dedicated to this problem. Such models are specifically designed to account for the ordered nature of the dependent variable without the need for postprocessing. These models incorporate the ordinal structure directly into their design, ensuring that the predictions respect the inherent ordering of the categories. Various approaches have been proposed in the literature to handle these constraints effectively.

One of the simplest and foundational methods is the Simple Approach to Ordinal Classification \cite{frank2001simple}. This method uses $n-1$, where $n$ is the number of ordinal classes. Each classifier predicts whether a given record belongs to a particular class or a higher one. This approach transforms the ordinal regression problem into a series of binary classification tasks, effectively modeling the ordinal nature of the dependent variable. The method of changing an ordinal regression model into a group of binary classification tasks is a common base for many methods that builds a more sophisticated idea on it \cite{bender1997ordinal, rennie2005loss, lechner2019orf}.

A group of methods uses linear models methods use linear models augmented with specialized loss functions tailored to ordinal data. For example, ORD \cite{bender1997ordinal} introduced an approach that adapts linear regression to ordinal problems, while Threshold-based models \cite{rennie2005loss} proposed a similar framework incorporating loss functions specifically designed to respect the order of categories.

Beyond linear models, more advanced approaches, such as those utilizing Gaussian Processes, have also been explored. Gaussian Process Ordinal Regression (GPOR) \cite{chu2005gaussian} method is a Gaussian process-based model for ordinal regression, using the probabilistic nature of Gaussian processes to capture uncertainty and the ordinal constraints of the problem.

Deep neural networks, with their ability to model complex, high-dimensional data, have been applied to ordinal regression tasks. Many neural approaches to ordinal regression also transform the problem into a set of binary classification tasks. Some of them fail to guarantee rank consistency, often assuming that the impact on model performance is negligible. However, this inconsistency can lead to illogical predictions, e.g., assigning a higher probability to class $i + 1$ than to class $i$ while still predicting class $i - 1$ as the most probable. Because of that problem, the assurance of enforcing non-decreasing was a subject of studies. For example, Consistent Rank Logits (CORAL) \cite{coral2020} introduced the loss function with guarantees to output ordinal consistency in the model outputs. Similarly, Conditional Ordinal Regression (CORN) \cite{shi2021deep} integrates ordinal constraints into the neural network training process. These innovations demonstrate the growing interest in the use of deep learning for ordinal regression.

\subsection{Evaluation}

To measure the performance of such algorithms, metrics such as the root mean squared error (RMSE) and the mean absolute error (MAE) are frequently used \cite{gutierrez2012experimental, saito2021evaluation, hu2018collaborative}. MAE is commonly utilized during the training process due to its robustness to outliers and its ability to provide a direct measure of average error. These metrics offer general perspectives on the performance of the model.

However, ordinal regression datasets are often characterized by class imbalance. To mitigate the risk that the majority classes disproportionately influence the evaluation, macroaveraged versions of MAE and RMSE are adopted, where errors are calculated per class and then averaged \cite{baccianella2009evaluation}. This approach ensures a more balanced assessment in all classes.

To evaluate how well a model preserves the ordinal relationships inherent in the data, Somers' D, a rank correlation metric specifically designed to capture ordinal dependence, is often used \cite{somers1962new}.

Given that ordinal regression can also be viewed as a form of classification, standard classification metrics such as accuracy are employed to measure the proportion of exact label matches between predictions and ground truth. Furthermore, due to the ordered nature of the target variable, the accuracy@k metric is used to assess how often the predicted label falls within a margin of k positions from the true label, thus accounting for near-miss predictions \cite{gaudette2009evaluation}.

\let\cleardoublepage\clearpage

\chapter{Methods}

This chapter presents the methods, algorithms, and models used to address the ordinal regression problem. It also provides a description of the characteristics of the dataset, the game itself, and the feature engineering techniques applied during preprocessing. 

\section{Dataset}
\label{section:methods_dataset}

Data were gathered from Pathfinder Roleplaying Second Edition rulebooks and textbooks. They consist of monsters represented by a set of all of their characteristics. Creatures are part of books that have been published over the years since the premiere of the second edition. 

Monsters are represented by multiple numerical attributes, such as strength = 10, fortitude = 35, or intelligence = –3. This numerical characterization naturally lends itself to a tabular data format, where each row corresponds to a monster, and each column to a specific attribute. Such structured data is well suited for machine learning tasks. Estimating the level of a monster, a categorical and ordered variable, based on those attributes can be formulated as an ordinal regression problem. Although comparing thousands of monsters across numerous parameters would be a highly demanding task for a human, it is a natural and efficient task for ML models.

\subsection{Feature engineering}

The following feature engineering methods were applied in the project:

\begin{itemize}
    \item creation of new features based on raw data that contain information relevant for the model,
    \item imputation of missing values using domain knowledge,
    \item feature normalization for models requiring such transformations (e.g. regularized regression).
\end{itemize}

Detailed implementations of these methods within the scope of this project are described later in this chapter.

The raw input data for our problem are complex and contain a significant amount of information that is irrelevant or not suitable for machine learning purposes. Each monster description consists of seven main fields, some of which are further divided into several, dozens, or even more nested subfields. In addition, certain parts of the data contain additional layers of complexity. The entire structure is represented as a deeply nested JSON dictionary, with a single file often spanning several hundred lines when displayed in a human-readable format.

Initially, a set of eight fundamental features was selected. These features define the core characteristics of each monster. 

\begin{enumerate}
    \item Strength (Str)
    \item Dexterity (Dex)
    \item Constitution (Con)
    \item Intelligence (Int)
    \item Wisdom (Wis)
    \item Charisma (Cha)
    \item Armor Class (AC)
    \item Hit Points (HP)
\end{enumerate}

The first six are standard attributes that form the basis of the system, and they are also used to derive many other statistics. The next two, AC and HP, represent defensive capabilities: AC determines how well the monster can avoid incoming attacks, while HP indicates how much damage it can take before dying.

This set was later extended with additional attributes that significantly influence monster difficulty level:

\begin{enumerate}
    \item Perception
    \item Fortitude
    \item Reflex
    \item Will
\end{enumerate}

Perception is a commonly used skill that helps a monster detect threats. Fortitude, Reflex, and Will are part of the so-called saving throws, which determine resistance to certain spells and effects. These are essential defensive metrics in the game system.

Other features used in the model:

\begin{enumerate}
    \item Speed: Land, Fly, Swim
    \item Number of Immunities
    \item Number of Spells per level (levels 1 to 9)
    \item Spell:
    \begin{itemize}
        \item Difficulty Class
        \item Attack
    \end{itemize}
    \item Melee attacks:
    \begin{itemize}
        \item Maximum attack bonus
        \item Average damage
    \end{itemize}
    \item Ranged attacks:
    \begin{itemize}
        \item Maximum attack bonus
        \item Average damage
    \end{itemize}
\end{enumerate}

Speed describes the movement capabilities of the monster across different types of terrain. If a monster cannot move in a given way (e.g., it cannot fly), the corresponding speed is set to 0.

Immunities represent complete resistance to specific types of damage. As these are relatively rare and diverse, only the total number of immunities was included as a characteristic, since treating them individually would not be statistically significant for model training and would result in numerous highly sparse features.

Some monsters have the ability to cast magic spells. Given the large number of spells available in the system, we include the number of spells that the monster can cast at each level, from 1 to 9. Higher-level spells are generally more powerful. Additionally, the highest spell level available to the monster is also included as a separate feature. Furthermore, the difficulty class (DC) of spells, which determines how challenging it is for a target to resist a spell cast by the monster, is also included. The spell attack modifier, used to resolve spell attacks against a target's AC, is incorporated similarly as a distinct attribute.

For modeling monster offensive capabilities, we considered both melee and ranged attacks. In the Pathfinder system, an attack involves two steps: a roll to hit using a 20-sided die (d20), subsequently modified by an attack bonus. If the hit is successful, a damage roll follows, using one or more dice plus a static modifier. For each monster, we extracted the highest available attack bonus and computed the expected average damage for the corresponding attack.

For example, for an attack described as \textit{"2d8 + 1d4 + 3"}, the expected damage is calculated as the sum of the expected values of each die roll plus the modifier:

$$2 \cdot \frac{1 + 2 + 3 + 4 + 5 + 6 + 7 + 8}{8} + \frac{1 + 2 + 3 + 4}{4} + 3= 14.5$$

\subsection{Dependent variable}

In the context of the problem addressed in this work, the dependent variable is the level of a monster, an integer value that ranges between -1 and 27 for Pathfinder 2e. Because all monsters above 20th level are extremely rare, they were aggregated into a single category, labeled level 21. The levels represent an ordinal scale, where higher values correspond to more challenging or powerful monsters.

\begin{figure}[htp]
    \centering
    \includegraphics[width=\textwidth]{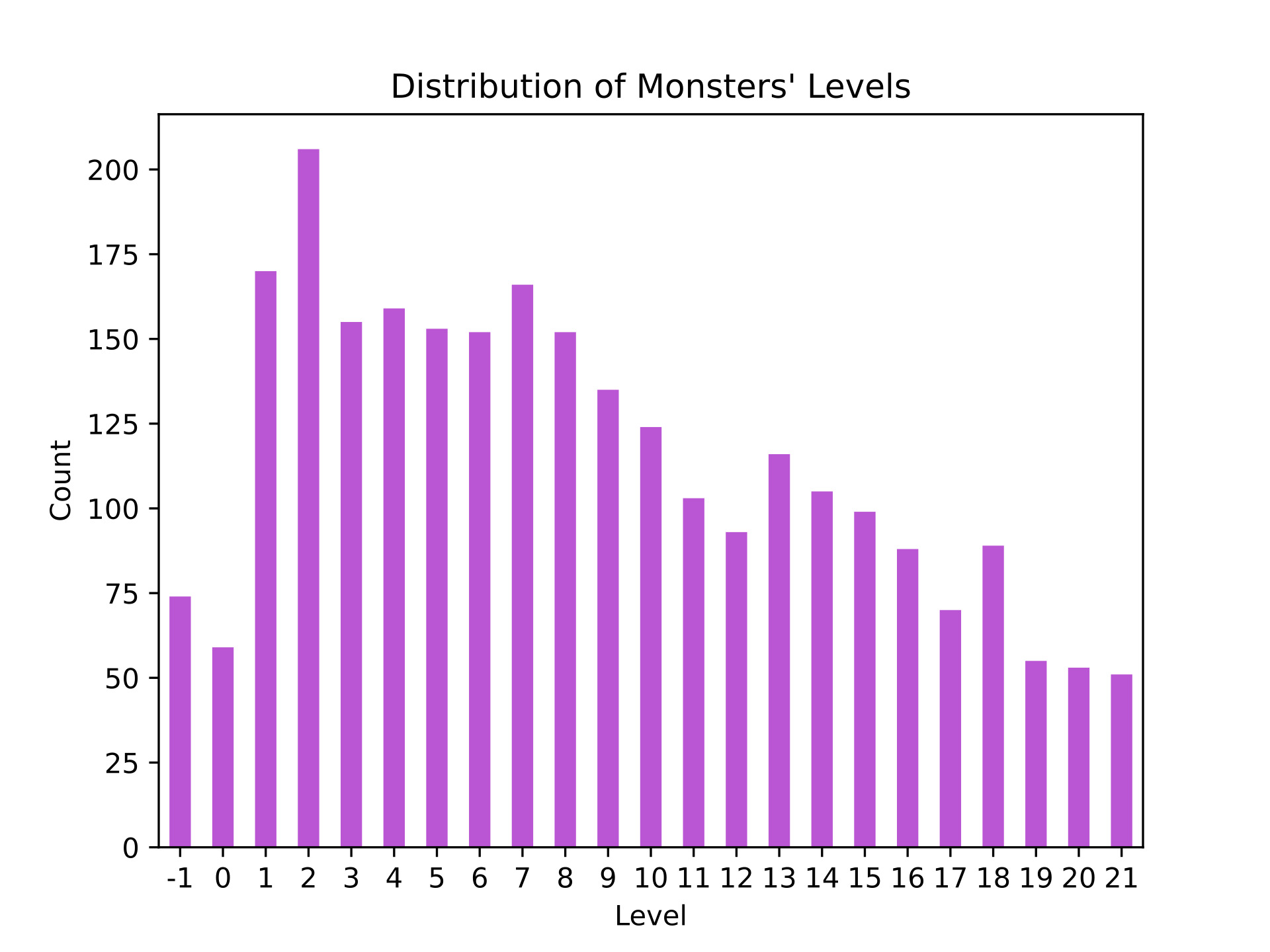}
    \caption{Accuracy}
    \label{fig:levels}
\end{figure}


\subsection{Pathfinder - Building creatures}

Creating custom monsters is a common practice in RPGs. The creators of Pathfinder have recognized this and included comprehensive guidelines for designing original creatures \cite{building-creatures}. The process differs significantly from the workflow applied with application of ML models. The information included in the Pathfinder Second Edition rulebook was used to design a baseline model that mimics the human approach to the problem.

The initial step involves conceptualizing the monster. One must consider whether the creature is, for example, more charismatic or more strength-oriented. Determining such a general monster type, called its archetype, is crucial, as it informs the distribution of its statistical attributes. For instance, a "soldier" archetype should feature high Strength, Fortitude, attack bonuses, and damage modifiers, along with high or extreme Armor Class (AC). Additionally, such monsters should possess tactical abilities, such as Reactive Strike.

The first statistic to define is the level of the monster. All other attributes are then determined relative to this level. Most statistics are categorized on a scale of low, moderate, high, and extreme in the context of monster level. Some attributes may even be assigned terrible values. It is essential to maintain internal balance. Allocating an extreme value to one attribute should be counterbalanced by assigning low or terrible values to others. For example, a monster with an extreme AC (hard to hit) might reasonably be given low hit points (HP) to preserve balance.

Based on the selected level and archetype, the final values should be chosen from the standardized skill and attribute tables, see Table \ref{table:skills}. These tables provide specific values or acceptable ranges for each attribute, depending on the level of the creature.

Whether to assign an extreme, high, moderate, or low value depends both on the monster's archetype and the desired power balance. The designer is responsible for specifying numerous characteristics and comparing the resulting monster to existing entries in the official bestiary to ensure it is well-balanced.

The collected monsters, tables, and domain knowledge were used to develop an algorithm to assign an archetype to a monster based on its characteristics. The level prediction is then based on the closest monsters that share the same archetype. This model is referred to as the baseline.

\begin{table}[H]
\centering
\begin{tabular}{ccccc}
\toprule
\textbf{Level} & \textbf{Extreme} & \textbf{High} & \textbf{Moderate} & \textbf{Low} \\
\midrule
--1 & +8   & +5   & +4   & +2 to +1    \\
0   & +9   & +6   & +5   & +3 to +2    \\
1   & +10  & +7   & +6   & +4 to +3    \\
2   & +11  & +8   & +7   & +5 to +4    \\
3   & +13  & +10  & +9   & +7 to +5    \\
4   & +15  & +12  & +10  & +8 to +7    \\
5   & +16  & +13  & +12  & +10 to +8   \\
6   & +18  & +15  & +13  & +11 to +9   \\
7   & +20  & +17  & +15  & +13 to +11  \\
8   & +21  & +18  & +16  & +14 to +12  \\
9   & +23  & +20  & +18  & +16 to +13  \\
10  & +25  & +22  & +19  & +17 to +15  \\
11  & +26  & +23  & +21  & +19 to +16  \\
12  & +28  & +25  & +22  & +20 to +17  \\
13  & +30  & +27  & +24  & +22 to +19  \\
14  & +31  & +28  & +25  & +23 to +20  \\
15  & +33  & +30  & +27  & +25 to +21  \\
16  & +35  & +32  & +28  & +26 to +23  \\
17  & +36  & +33  & +30  & +28 to +24  \\
18  & +38  & +35  & +31  & +29 to +25  \\
19  & +40  & +37  & +33  & +31 to +27  \\
20  & +41  & +38  & +34  & +32 to +28  \\
21  & +43  & +40  & +36  & +34 to +29  \\
22  & +45  & +42  & +37  & +35 to +31  \\
23  & +46  & +43  & +38  & +36 to +32  \\
24  & +48  & +45  & +40  & +38 to +33  \\
\bottomrule
\end{tabular}
\caption{Skills}
\label{table:skills}
\end{table}

\section{Classical regression}

The most basic approach to ordinal regression is directly using regular regression models. However, there is an ongoing discussion on whether such variables can be treated as continuous \cite{winship1984regression}. Despite the development of many dedicated algorithms specifically designed for ordinal regression, the approach of using continuous predictions followed by a mapping step has remained a subject of research \cite{kramer2001prediction, shin2022moving, regression-to-build-classifiers}. The main motivation behind this approach is the availability of a wide range of well-established and high-performing models for continuous regression, which can be used easily without modification. This method requires two components: the regression model and method of rounding results to ordinal values, i.e. mapping real numbers to integers.

\subsection{Regression models}

As part of this work, the most popular models for traditional continuous regression were used. The whole set was introduced to see not only a difference between classical models and dedicated ordinal models but also among different models from each group after hyperparameters tuning.

\subsubsection{Linear Models}
A set of linear models with different loss functions and regularization methods was used. LASSO, Ridge, LAD, and Huber regression.
Linear models for regression constitute a baseline model for the traditional regression group. Those models assume a linear relationship between the features (independent variables) and the dependent variable in the following form:

\begin{equation}
\hat{y} = w_{0} + w_{1}x_{1} +... + w_{d}x_{d} = x^{T}w
\end{equation}
where $d$ is the number of features.

Linear models share the same functional form, yet differ in their choice of loss function and regularization methods. This results in different weight vectors $w$ and regressor properties.

In case of a linear regression model, the model parameters $w$ are determined by minimizing the mean squared error (MSE), expressed by the formula:
\begin{equation}
L_{MSE} = \sum_{i=1}^{n} \left(y_i - \hat{y}_i \right)^2,
\end{equation}
where $n$ is the number of samples.

Models with regularization L1 (LASSO) and regularization L2 (Ridge) were used, tuning the regularization strength $\lambda$. The corresponding formulas for both types of regularization are given below. 

\begin{equation}L_{LASSO}(w) = \sum_{i=1}^{n} \left(y_i - \hat{y}_{i}^Tw \right)^2 + \lambda  \sum_{j=1}^{d} \left| w_{j} \right| 
\end{equation}

\begin{equation}
L_{Ridge}(w) = \sum_{i=1}^{n} \left(y_i - x_{i}^Tw \right)^2 + \lambda  \sum_{j=1}^{d}w_{j}^2
\end{equation}

Ridge regression tends to produce more stable models by discouraging large weight values, thereby improving the robustness to overfitting. In contrast, LASSO not only shrinks the coefficients, but also marks some of them to exactly zero, effectively performing feature selection. This results in simpler models that are easier to interpret.

Least Absolute Deviations (LAD) regression optimizes the mean absolute error (MAE) as the loss function. This model uses only L1 regularization (omitted in the formula for brevity) and is robust to outliers.

\begin{equation}
    L_{LAD}(w) = \sum_{i=1}^n |y_i - \hat{y}_i|
\end{equation}

In practical applications, the parameters of the LAD model are typically estimated using linear programming techniques.

Lastly, Huber regression is based on a Huber loss, a type of loss function designed for robust regression, particularly when data may contain outliers. Similarly to LAD regression, it is less sensitive to outliers than the Mean Squared Error (MSE) loss. However, Huber loss combines the strengths of both MSE and Mean Absolute Error (MAE) by applying a different metric depending on the value of the residuals. A major advantage of Huber regression is that it is faster than LAD regression.  For small residuals, the Huber loss behaves like MSE, ensuring smooth gradients and stable convergence. For large residuals, it switches to behave as MAE, reducing the influence of outliers. The transition point between these two behaviors is controlled by a hyperparameter $\epsilon$. One of the practical advantages is that its performance is not overly sensitive to the precise value of $\epsilon$, making it relatively easy to adjust. It is typically used with L2 regularization.

\begin{equation}
L_{Huber}(w) = 
\begin{cases}
\frac{1}{2} ||y-Xw||^2 & \text{if } ||y-Xw|| \leq \epsilon\\
\epsilon \cdot (||y-Xw||_1-\frac{1}{2} \epsilon) & \text{otherwise }
\end{cases}
\end{equation}

\subsubsection{Random Forest}

Random Forest (RF) \cite{breiman2001random} is a type of ensemble learning model that combines multiple decision trees to produce more stable and accurate predictions. RF builds multiple decision trees using bootstrap samples generated from the input training dataset. Each tree is trained independently on a different sample, with a random subset of selected features (random subspace method). During prediction, their outputs are averaged. This procedure reduces the variance of the model and results in a model relatively insensitive to hyperparameters \cite{rf-hyperparameters}. In regression, the Mean Squared Error (MSE) is used as a splitting criterion during tree construction.

\subsubsection{LightGBM}

LightGBM \cite{lightgbm-def} is an ensemble learning algorithm and a particularly efficient implementation of gradient boosting. It uses multiple decision trees that grow leaf-wise \cite{xgboost}, unlike, for example, the XGBoost algorithm. This means that nodes are added only in regions that provide the greatest reduction in the cost function. This allows for fast and effective training that provides high prediction accuracy. LightGBM also stands out for its speed and scalability to large datasets. A drawback of the algorithm is the large number of hyperparameters, which requires sophisticated optimization algorithms.

In this case, the Tree Parzen Estimator (TPE) \cite{tpe} was used. It is an efficient Bayesian hyperparameter optimization algorithm implemented in the Optuna library \cite{akiba2019optuna}. This algorithm focuses on efficiently exploring the hyperparameter space to find a configuration that minimizes the cost function. Unlike a random hyperparameter search, the algorithm uses Bayesian inference to determine how specific hyperparameters affect the model, to inform the hyperparameter search.

\subsubsection{SVM}

Support Vector Machines (SVMs) are a class of supervised learning algorithms widely used for classification and regression tasks. The fundamental idea behind SVMs is to transform the input data into a higher-dimensional feature space using a nonlinear mapping, where a linear decision boundary can then be constructed \cite{cortes1995support}. This approach allows the model to handle complex, non-linear relationships in the original input space.

SVMs are known as maximum-margin classifiers, which means they aim to find the hyperplane that maximally separates data points from different classes. This characteristic contributes to their robustness against overfitting and resilience to noisy data. Furthermore, SVMs are particularly effective in high-dimensional spaces.

Another strength of SVMs lies in their flexibility through the use of kernel functions, which enable the algorithm to implicitly perform the non-linear transformation without explicitly computing the high-dimensional feature space. Common kernel functions include, for example, linear and radial basis function (RBF), which were used for the problem presented.

Linear:
\begin{equation}
    K(x_i, x_j) = x_i \cdot x_j
\end{equation}
A particular variant of the RBF kernel was used, a Gaussian kernel. In this case, $\gamma$ is inversely proportional to the variance of the input data.
\begin{equation}
K(x_i, x_j) = e^{-\gamma {||x_i-x_j||^2}} 
\end{equation}

\subsubsection{K-Nearest Neighbours}

K-Nearest Neighbors (KNN) is a non-parametric supervised learning algorithm commonly used for both classification and regression tasks. KNN operates as a proximity-based classifier, where an unseen data point is assigned to a class based on the majority vote of its $k$ nearest neighbors in the feature space.

In the case of regression, the algorithm adopts an analogous approach: instead of voting, it predicts the output as the average of the values corresponding to the k nearest neighbors, so for $k=1$, the prediction is simply the value of the closest training point.

KNN can employ various distance metrics to determine the proximity of neighbors, including the Euclidean, Manhattan, and Minkowski distances, among others. The basic implementation of KNN assigns uniform weights to all k nearest neighbors, treating each neighbor equally in the prediction. However, in practice, it is often advantageous to apply distance-weighted voting, where closer neighbors exert a stronger influence on the final prediction. A common approach is to assign weights inversely proportional to the distance from the query point, giving greater importance to nearby observations.

\subsection{Rounding}

When continuous regression models are used for ordinal prediction tasks, an additional step is required to map the continuous output to discrete ordinal labels. Some studies propose to perform this mapping through an additional classification model to determine the appropriate discrete class \cite{regression-to-build-classifiers}, while others employ simpler techniques, such as mathematical rounding with threshold 0.5 \cite{shin2022moving}. In this project, the latter approach was used as a baseline. In addition, several alternative strategies for the best threshold finding techniques were explored:

\begin{itemize}
    \item Mathematical rounding
    \item Tuned global threshold
    \item Tuned thresholds per level
    \begin{itemize}
        \item Treated as hyperparameter
        \item Mathematical programming formulated as a graph shortest paths problem
    \end{itemize}
\end{itemize}

In case of labels $1, 2, \dots, n$, there are $n - 1$ thresholds, denoted as $t_1, t_2, ..., t_i, t_{i+1}, ..., t_{n-1}$, where $t_i$ is a threshold between level number $i$ and $i + 1$ and its value is between $i$ and $i + 1$ and satisfies $i < t_i < i + 1$. 
The mapping to level for a given threshold $t_i$ is defined as:

\begin{equation}
     x \rightarrow \begin{cases}
        i &   i \leq x < t_i \\
        i + 1 & t_i \leq x < i + 1
    \end{cases}
\end{equation}

Given a loss function $L$, the objective is to find a set of thresholds such that the rounding procedure described above minimalizes the cost computed by $L$.

Mathematical rounding is a widely used method in ordinal regression settings with continuous models \cite{kramer2001prediction, shin2022moving}. This rounding uses a fixed threshold of 0.5. For example, considering levels ranging from -1 to 21, the mapping is as follows: $(-1.5; -0.5)$ rounds to -1, $<-.5; 0.5)$ rounds to 0, ... and $<20.5; 21.5)$ rounds to 21. Mathematical rounding is the only method in this project that is applied without any optimization. However, there is a possibility that the fitted model may slightly overestimate values for some levels, underestimate values for another set of levels, and produce almost perfect estimations for yet another group. These discrepancies can arise due to an imbalanced regression present within the dataset, see Figure \ref{fig:levels}. In such cases, utilizing a set of thresholds distinct from the classical ones may give better results. All other methods aim to improve performance by optimizing the rounding thresholds.

The single best threshold strategy was initially proposed based on manual experimentation in a previous thesis \cite{rpg-inz}, where different threshold values were tested for the models used. In the current work, this process has been automated: a range of thresholds is evaluated after model training, and the one with the best performance is selected for prediction. Unlike mathematical rounding, which uses a fixed midpoint threshold of 0.5, this method applies a threshold $t$ that yields the best performance. Each real-valued (continuous) prediction is rounded according to the following, see Equation \ref{equation:single_threshold}. 
\begin{equation}
     x \rightarrow \begin{cases}
        i &   i \leq x < i + t \\
        i + 1 & i + t \leq x < i + 1
    \end{cases}
    \label{equation:single_threshold}
\end{equation}

However, here we have the problem of an imbalanced classification. There are considerably more low-level monsters than high-level ones, which makes a single threshold not adaptive enough for different parts of the data.
This observation led to the introduction of the best threshold per-level strategies. Two methods were investigated for this purpose. The first explicitly treats all 20 thresholds (in case of Pathfinder) as hyperparameters. Their values are determined using the Tree Parzen Estimator (TPE) algorithm of Optuna \cite{tpesampler, tpe}, to find the optimal thresholds between each pair of neighboring ordinal levels. The second approach, originally proposed in this work, involves formulating the threshold selection problem as a linear programming task. It can be reformulated and implemented in practice as the shortest path search on a graph, enabling a quick and deterministic identification of the best combination of thresholds across all levels. Graph rounding was proposed and implemented to select thresholds more efficiently than the TPE method and to produce deterministic results, unlike the Optuna multithreshold rounding method. 

To introduce graph-based rounding, we need to formally define the problem of optimal rounding in the context of ordinal regression.
There are two sets of values: ordinal ground truth classes (integers) $y$, and real values returned by regression $\hat{r}$. Denote the real values returned by the regression model as $\hat{r}$. We need to round them to integers, which are the ordinal predictions of the model, denoted $\hat{y}$.  In case of Pathfinder Second Edition, they are from the set $\{-1, 0, 1, ..., 20, 21\}$. Assume that we optimize the sum of absolute deviations between predicted and true labels, $\sum_{j=1}^n|y_j - \hat{y}_j|$ for $n$ observations, by optimizing the intervals of rounding thresholds $[a_i, a_{i+1}]$ for which the number is rounded to $i$. For each $\hat{r}_j$, there is $\hat{y}_j = min_i(|\hat{r}_j - \frac{a_{i+1} - a_i}{2}|)$ i.e. the threshold between two integers that optimally rounds prediction (in terms of MAE). It can be formulated as follows:

\begin{gather*}
    \text{minimize } \sum_{j=1}^n|y_j - \hat{y}_j| \\
    \text{for variables } a_i, i= -1, 0, \dots, 21 \\
    \text{where } \hat{y}_j = \text{arg } min_i|\hat{r}_j - \frac{a_{i+1} - a_i}{2}| \\
    \text{subject to} \\
    a_{-1} = -1 \\
    a_{21} = 21 \\
    a_i < a_{i+1}, i = -1, 1, \dots, 20
\end{gather*}

This can be reformulated as a shortest path problem in a directed, weighted graph, organized conceptually in layers. Let $R$ be a set of distinct values $r$ that symbolize thresholds. This set is manually defined before optimization. In the experiment, two sets of were tested:
$R_1=[0.05, 0.1, 0.15, \dots, 0.95]$ and $R_2=[0.25, 0.3, 0.35, \dots, 0.75]$. The set of ones used for building the graph is given at the start.  Let $I=\{1, ..., n - 2\}$ for the level values $=\{0, ..., n-1\}$. Graph vertices (nodes) are created in the lattice $R \times I$, thus represented as pairs $(r, i)$. They are connected with directed edges (arcs), which go from $(r, i)$ to $(r', i+1)$, where $r < r'$. An edge $(r, i) \rightarrow (r', i+1)$ is traversed if all observations with $r \leq r_j \leq r'$ have $\hat{y}_j = i$. The cost of traversing the arc is the sum of absolute deviations:

\begin{equation}
    \sum_{j: r \leq r_j \leq r'} |y_j - i|
\end{equation}

\begin{figure}[H]
    \centering
    \includegraphics[width=0.75\linewidth]{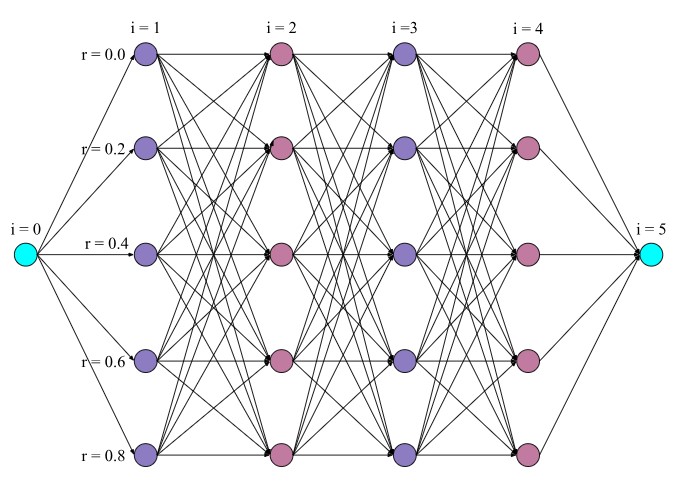}
    \caption{Rounding graph}
    \label{fig:graph}
\end{figure}

There is also a source node $(0, -1) $ with arcs to all $(r, 0)$ nodes and a sink node with arcs from all $(r, 20)$. The shortest path from source to sink will result in the best route with given $r$, where thresholds $a_i$ take the chosen $r$ as values. In this way, the created path defines the thresholds that minimize the overall loss function, achieving the same goal as linear programming. The distinct advantages of this approach are the simplicity of implementation and low computational cost. An example graph with five levels and thresholds $R = \{0.0; 0.2; 0.4; 0.6; 0.8\}$ is shown in Figure \ref{fig:graph}.

\section{Ordinal regression methods}
\label{sec:ordinal_regression_methods}

The application of classical regression algorithms to ordinal regression problems has two drawbacks. First, it restricts the possibilities to regression models only, which have inherent limitations, e.g. they cannot account for imbalanced classes in ordinal regression. Second, it raises the challenge of how to reliably map continuous predictions back to discrete ordered classes. This section describes algorithms designed specifically for the ordinal regression problem, which is also called ordinal classification. They are in a way classifier-regressor hybrids that take advantage of the ordinal structure of the target variable.

\subsection{A simple approach to ordinal classification (SAOC)}

An intuitive and simple approach to ordinal classification (SAOC) enables the use of standard classification algorithms for ordinal regression tasks without requiring modifications to existing classifier implementations \cite{frank2001simple}. This method leverages the ordinal structure of the target variable through the transformation of the original labels.

\begin{figure}
    \centering
    \includegraphics[width=0.75\linewidth]{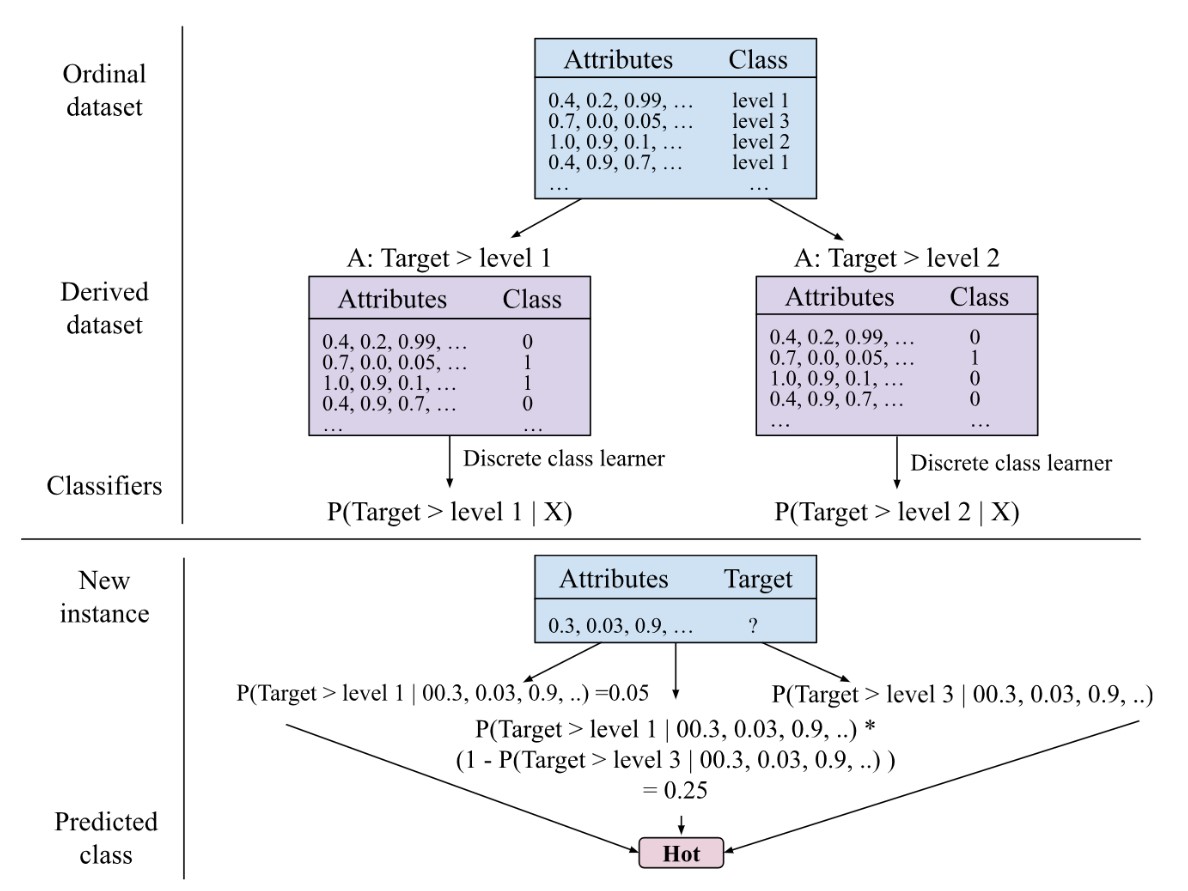}
    \caption{Application of standard classifiers to ordinal regression.}
    \label{fig:sor}
\end{figure}

The ordinal problem with $k$ classes is decomposed into $k-1$ binary classification problems, see Figure \ref{fig:sor}.  
Each of the binary classifiers $\{1, \dots, k-1\}$ is trained to predict whether the class is equal to or greater than a certain level. In other words, $i$-th classifier predicts the probability that class is larger than $i$. For the $i$-th classifier, the dataset is transformed using the following rule:

\begin{equation}
y' =
\begin{cases}
    1 & \text{if } y \leq i \\
    0 & \text{if } y > i \\
\end{cases}
\end{equation}

This transformation results in $k-1$ binary datasets, each of which contains the same input features but different binary labels, according to the threshold applied. Each dataset is then used to train a separate binary classifier, treating the task as a standard binary classification problem.

To predict a value for a given record, each classifier is first used to calculate the probability $P(Target>V_{i})$. Then those outputs are used to receive the probabilities of each class. The probability of the highest ordinal value is $P(Target > V_{k-1})$ and the probability of the first class is given by $1 - P(Target > V_1)$. For middle classes, the probability $P(V_i)$ depends on a pair of classifiers. The classifiers' probabilities are transformed into the final probabilities of classes using the following equations:

\begin{equation}
P(V_i)=
\begin{cases}
    1-P(Target > V_1) & \text{if } i = 1 \\
    P(Target > V_{i-1}) * (1-P(Target > V_i)) & \text{if } 1 < i < k \\
    P(Target > V_{k-1}) & \text{i = k} \\
\end{cases}
\end{equation}

The class with the highest probability is then assigned to the given record. There is no additional step to ensure the consistency of the predictions.

\subsection{Ordered Logistic Regression (ORD)}

One of the most widely used models for binary classification is logistic regression. It is based on the idea of modeling the probability of a class using a linear function of the input features, transformed with the logistic (sigmoid) function. Logistic regression assumes a linear relationship between the independent variables and the logit, i.e. the logarithm of the odds.

In the standard multinomial logistic regression, the model estimates the log-odds of each class $i$ relative to a reference class $k$ using probabilities. The probability for the category $i$ is given by $P(Y = i) = p_i \text{ for i=1, ...,k}$. Logits are calculated as follows:

\begin{equation}
    logit(Y=i | x)= \ln{\frac{p_i}{p_{k}}},  \text{ for i=1, ...,k - 1}
\end{equation}
\begin{equation}
    logit(Y=i |x )=b+w_{1}x_1+...+w_{d}x_d, \text{ for i=1, ...,k - 1}
\end{equation}

However, in the case of ordinal dependent variables, the standard multinomial logistic regression does not use the information regarding ordering between categories. To address this, Ordered Logistic (ORD) was proposed \cite{bender1997ordinal}. The main idea is to introduce class ordering information into the model, quite similarly to the simple approach to ordinal regression (described in the previous section). Instead of directly modeling the probabilities of individual classes, this approach models the cumulative probabilities of the outcome being less than or equal to a given class $V_i$. The main difference is that ORD uses a consistent and cumulative probability instead of treating each class separately, especially during the final prediction step.

The cumulative probability of the result belonging to class $i$ or below is given by:

\begin{equation}
    P(Y \leq V_i) = p_1+...+p_i
\end{equation}

The corresponding odds and logit of this cumulative probability are as follows:

\begin{equation}
    odds(Y \leq i) = \frac{P(Y \leq i)}{P(Y > i)} = \frac{p_1 + ... + p_i}{p_{i + 1} + ... + p_{k-1}}
\end{equation}
\begin{equation}
    logit(Y \leq i) = \ln{\frac{P(Y \leq i)}{1 - P(Y \leq i)}}, \text{ i =1, ..., k - 1}
\end{equation}

This model is based on the proportional odds assumption, which states that the effect of predictors is constant across all thresholds. In other words, if a particular feature increases the probability that an observation belongs to class $i$, it does so consistently across all boundaries between classes. The regression coefficients $w$ remain the same for each threshold, while only the intercepts $b_i$ vary.

\subsection{Threshold-based models}

Threshold-based models generalize classical binary classification approaches (e.g. logistic regression) to ordinal regression by extending the concept of thresholds to multiple ordered categories \cite{rennie2005loss}. Specifically, the method introduces $k-1$ thresholds $\theta_1 < \theta_2< ...<\theta_{k-1}$, which partition the real line into $k$ segments. The exterior segments are set as infinite and are denoted as $\theta_0 = - \infty$ and $\theta_k = +\infty$. Each segment corresponds to one of the $k$ ordered classes. A predictor score $z$ that is in the range $\theta_{i-1} < z < \theta_i$ results in the assignment of the instance to the $i$-th class. This framework generalizes the single-threshold setting used in binary classification.

There are two variants of loss functions used in this approach, as described below. Both constructions can be used to generalize the loss function $f$ for binary labels, e.g. binary cross-entopy or hinge loss. They are designed to penalize predictions that fall outside the correct threshold-defined interval but differ in how they handle and interpret the severity of the misclassification. Each threshold violation is penalized using the given loss function $f$. In this work, hinge loss was always used for this purpose:

\begin{equation}
    L_h(y, \hat{y}) = max(0, 1 - y * \hat{y})
\end{equation}

\subsubsection{Immediate-Threshold}

The Immediate-Threshold loss focuses on thresholds that directly define the correct class for a given instance. For an example labeled $(x, y)$, where y is part of $i$-th class, only two thresholds define the range $(\theta_{i-1}; \theta_i)$ within which the answer is considered acceptable. The predictor output for $x$ is denoted as $z = z(x)$. All violations of these values are penalized:
\begin{equation}
    L(z, i)=f(z-\theta_{i-1}) + f(\theta_i-z)
\end{equation} 

This formulation ensures that no penalty is incurred if the predicted value falls within the correct interval $(\theta_{i-1}; \theta_i)$. However, if the prediction lies outside this range, the penalty of $f$ discourages the error. Compared to the standard version of $f$, e.g. hinge loss, this definition results in a symmetric loss function that applies $f$ to violations beyond each boundary of the correct interval.

It is worth noticing that if $f$ is an upper bound for zero-one error in the case of binary classification, then the Immediate-Threshold loss is an upper bound on the zero-one loss for ordinal regression. That is, it guarantees no loss for correct predictions and applies penalties only for threshold violations, regardless of how many thresholds are crossed.

\subsubsection{All-Threshold}

The Immediate-Threshold loss is sensitive to whether a prediction lies within the correct interval. However, it does not consider the magnitude of the error, that is, how far the prediction is from the true class.  It fails to use the actual ordering of the classes, leading to poorer performance on metrics such as mean absolute error (MAE) that explicitly measure this.

To address this, the All-Threshold loss introduces a formulation that sums penalties for violations of all thresholds. In this way, it encourages not only correct classification, but also making predictions that are closer to the target class. It is defined as:

\begin{equation}
    L(z,y) = \sum\limits_{l=1}^{K-1}
    \begin{cases}
        f(\theta_l - z) & \text{if } l \geq y \\
        f(z - \theta_l)& \text{if } l < y
    \end{cases}
\end{equation}

This formulation penalizes the model each time a threshold is crossed, and the magnitude of the loss increases linearly with the distance between the predicted and the correct class. This aligns more directly with the objectives of ordinal regression, such as MAE minimization.

\subsubsection{Learning thresholds}

Training ordinal regression models using threshold-based losses requires optimizing both the predictor parameters (e.g., weights in linear models) and the threshold values $\theta$. Unlike fixed, uniformly spaced thresholds, learning thresholds as part of model training allow the algorithm to adapt to the true distribution and usage of labels. This approach addresses the problem of imbalanced regression regression and allows better alignment between the model outputs and the structure of the ordinal label space.

\subsection{Gaussian Processes for Ordinal Regression}

Gaussian processes (GPs) provide a kernel-based probabilistic approach to ordinal regression. In this method, the latent function $K$ is modeled using a Gaussian process, and the likelihood function is defined using a threshold model that generalizes the probit function for ordinal targets. This model interprets the ordinal label as an interval within the continuous latent function space defined by the thresholds.

To model ordinal output, a set of thresholds $\{\theta_0,\theta_1,  ..., \theta_{K-1}\}$ is used to partition the latent function output $F$ into ordered categories $K$. These thresholds define the boundaries between adjacent ordinal classes. Given a latent value $F$,  the likelihood that an observation belongs to a particular class is expressed using the standard normal cumulative distribution function $\phi(\cdot)$ (i.e., the probit function), see equation \ref{equation:gp_function}. Here, $\theta$ are the threshold parameters and $\sigma$ is a learned scaling parameter (noise).

\begin{equation}
\begin{aligned}
P(Y = 0 | F) & =  \phi(\frac{\theta_0 - F} {\sigma}) P(Y = 1 | F) \\
 & =\phi(\frac{\theta_1 - F} {\sigma}) - \phi(\frac{\theta_0 - F} {\sigma}) P(Y = 2 | F)  = \dots \\
 & = 1 - \phi(\frac{\theta_{K-1} - F} {\sigma}),
 \end{aligned}
\label{equation:gp_function}
\end{equation}

The parameters $\{\theta_0, ..., \theta_{K-1}\}$  and $\sigma$ are learned by maximizing the posterior probability given the observed data $D$:

\begin{equation}
    P(\theta, \sigma | D) \propto P(D | \theta, \sigma)P(\theta),
\end{equation}

where $P(D|\theta,\sigma)$  is the likelihood of the data under the current parameters and prior is specified using domain knowledge, or alternatively by using an uninformative prior. In this work, an uninformative prior was used.

This model leverages the flexibility of Gaussian processes while explicitly modeling the ordinal structure of the output via a cumulative likelihood. A known limitation of this approach is its poor scalability with respect to the number of training samples, on the order of $O(n^3)$. In the case of the Pathfinder, this is not particularly problematic, as the number of monsters is relatively small, around 2600.

\subsection{Ordered Random Forest}

The Ordered Random Forest (ORF) is a machine learning estimator designed for ordered outcome variables based on the Random Forest algorithm \cite{lechner2019orf, lechner2025random}. Its key advantage lies in its ability to handle high-dimensional covariate spaces efficiently. The main objective of the ORF is to estimate the probabilities of conditional choice $P(Y_i=k | X_i =x)$.


The ordered choice probabilities are constructed from cumulative, nested binary indicators. Specifically, binary indicators are defined as $Y_{k,i} = \mathbb{1}_{(Y_i \leq k)}$ for classes $k \in \{1, \dots, K - 1\}$. The model first estimates $K - 1$ binary classification problems using Random Forests, producing cumulative probability estimates:

\begin{equation}
    \hat{Y}_{k,i} = P(Y_{k,i}=1 | X_i =x)
\end{equation}

From these cumulative estimates, the predicted probabilities for each ordinal class are derived as follows:

\begin{equation}
    \hat{P}_{k,i} = \hat{Y}_{k,i} \text{ if } k = 1
\end{equation}

\begin{equation}\label{equation:orf-probs}
    \hat{P}_{k,i} = \hat{Y}_{k,i}  - \hat{Y}_{k - 1, i} \text{ if } 1 < k \leq K
\end{equation}

\begin{equation}
    \hat{Y}_{K,i} = 1
\end{equation}

\begin{equation}\label{equation:orf-non-neg-probs}
    \hat{P}_{k,i} = 0 \text{ if } \hat{P}_{k,i} < 0
\end{equation}

\begin{equation}\label{equation:orf-normalize}
    \hat{P}_{k,i} = \frac{\hat{P}_{k,i}}{\sum_{k=1}^{K} \hat{P}_{m,i}} \text{ for } 1 \leq k \leq K
\end{equation}

Equation \ref{equation:orf-non-neg-probs} ensures that the class probabilities do not take negative values, which may occur if certain categories have very few observations. Although this is theoretically possible, it is rare in practice. If any probabilities are set to zero, Equation \ref{equation:orf-normalize} renormalizes the vector of probabilities so that they sum to one.

\subsection{Deep Neural Networks for Ordinal Regression}

In this work, some representative neural network models for ordinal data have been used. Those models are defined primarily on the basis of their loss functions, incorporating inductive bias appropriate for ordinal regression. All methods were applied using the same simple multilayer perceptron (MLP) architecture, which was then adapted to each specific approach, primarily by modifying the dimensionality of the output layer, depending on the requirements of the given method. both methods addressing the rank inconsistency problem that is an issue for methods like NNRank or OR-CNN.

\subsubsection{Spacecutter}

Spacecutter \cite{spacecutter} is an ordinal regression technique designed to learn how to partition the prediction space using cutpoints, similarly to threshold-based methods. The core idea is to determine the optimal positions for these cutpoints that separate ordered classes. This approach builds upon binary logistic regression and extends its probabilistic interpretation to the ordinal setting for neural networks.

In binary logistic regression, the point where the predicted logit value is equal to zero corresponds to a 50\% probability of belonging to either class, serving as the decision boundary. Spacecutter generalizes this by introducing $K-1$ learnable cutpoints for $K$ ordinal classes, enabling the model to segment the prediction space into $K$ intervals, each representing a class.

Let $f(X)$ denote the output of the model (e.g. a linear function of input features), and let $\{c_0, ..., c_{K-1}\}$ be the set of learned cutpoints, where $c_0<c_1<...<c_{K-1}$.  The cumulative logistic link function is then used to model the probability that a given observation belongs to a specific class $k$:

\begin{equation}
P(y = k) =
\begin{cases}
    \sigma(c_0-f(X)) & \text{if k = 0} \\
    \sigma(c_k - f(X)) - \sigma(c_{k-1} - f(X)) & \text{if 0 < k < K} \\
    1 - \sigma(c_{K-1} - f(X)) & \text{if k = K}   
\end{cases}
\end{equation}

For K = 2, this formulation reduces to the standard binary logistic regression model. In practice, both the model parameters (e.g., weights in a linear model) and the cutpoints $\{c_k\}$ are learned jointly during training.

The Spacecutter model is trained by minimizing the negative log-likelihood (NLL) of the observed labels under the predictive distribution of the model. Given a dataset: $\{(x_i, y_i)\}_{i=1}^n$, the loss function is defined as:

\begin{equation}
L=-\sum\limits_{i=1}^{n} logP(y_i = k | x_i),
\end{equation}

where $P(y_i = k | x_i)$ is computed using the cumulative logistic formula shown above. This objective encourages the model to produce probabilities that are well-calibrated and consistent with the ordinal nature of the labels. Because cutpoints are learned directly from the data, Spacecutter can flexibly adapt to label distributions and spacing that may not be uniform.

\subsubsection{NNRank}

The NNRank model \cite{cheng2008neural} transforms a standard neural classification architecture trained with cross-entropy into an ordinal regression model by modifying the label encoding and output interpretation. Instead of generalizing binary classification, the model starts from the idea of classification but with subcategories. Each point that belongs to rank $i$ also belongs to rank $\{1, \dots, i-1\}$. For a problem with $K$ ordered classes, each target label is transformed into a binary encoded vector of size $K$  to reflect the ordinal structure. Specifically, for a label $k \in \{1, ..K\}$, the corresponding target vector is defined as:

\begin{equation}
    class_k = [\text{1 if i } \leq \text{ k else 0] for }i\in \{1, ..K\}
\end{equation}

For example, in a 5-class ordinal regression problem, the labels would be encoded as:

$$class_1 = [1, 0, 0, 0, 0]$$
$$class_2 = [1, 1, 0, 0, 0]$$
$$class_3 = [1, 1, 1, 0, 0]$$
$$class_4 = [1, 1, 1, 1, 0]$$
$$class_5 = [1, 1, 1, 1, 1]$$

This encoding preserves the ordinal relationships between classes: the distance (e.g. Hamming distance) between $class_i$ and $class_{i+2}$ is greater than between $class_i$ and $class_{i+1}$, reflecting their ordering.
The model ends with a sigmoid-activated output layer a $K$-dimensional vector $\hat{y} = [\hat{y}_1, ..., \hat{y}_K]$,  where $i$-th element represents the probability that the class is greater or equal to $i$.

The network is trained using a binary classification loss in each of the $K$ outputs independently. Two common loss functions are binary cross-entropy (BCE) or mean squared error. For this thesis, the former was used.

During inference, the predicted class is determined by applying a threshold of 0.5 to each predicted probability in $\hat{y}$ and counting the number of values greater than 0.5 from left to right until the output of the node is smaller than the threshold $0.5$ or there are no nodes left.

One known issue is the possibility of rank inconsistency, where the predicted vector contains a ``dip'' in the expected monotonic pattern, for example [0.88, 0.9, 0.44, 0.71, 0.3]. Such cases could ambiguously be assigned to class 2 or class 4. However, in empirical evaluations, such inconsistencies did not appear and did not affect model performance.

\subsubsection{OR-CNN}

OR-CNN \cite{niu2016ordinal} is a loss function used for ordinal regression, particularly in CNNs for age estimation. It reformulates the ordinal regression problem as a series of binary classification problems. Specifically, an ordinal problem with $K$ classes is transformed into $K - 1$ binary classification tasks. This approach is similar to NNRank. Classes are encoded analogously, but the decoding is different. The final prediction for an unseen record $X$ is obtained by summing the outputs of the $K - 1$ classifiers.

\begin{equation}
k = 1 + \sum \limits_{i = 1}^{K - 1} f_k(X),
\end{equation}

Here, $f_k(X) \in \{0, 1\}$ is the binary output of the $i-th$ classifier for input $X$. No consistency constraints are enforced between classifiers, which simplifies training.

The model is trained using a weighted binary cross-entropy loss. Let $\lambda_k$ denote the importance coefficient of the binary task $k-th$. The weights $\lambda_i$ are calculated based on the training data distribution:

\begin{equation}
    \lambda_i = \frac{\sqrt{N_k}}{\sum \limits_{k = 1}^{K} \sqrt{N_k}},
\end{equation}

where $N_k$  is the number of samples in class $class_k$, and the denominator normalizes the weights. This scheme gives higher weight to classifiers with thresholds near densely populated class regions, improving their training effectiveness. This approach is particularly beneficial in the context of this work, as the Pathfinder dataset presents an example of an imbalanced regression problem.

\subsubsection{Consistent Rank Logits (CORAL)}

CORAL \cite{coral2020} addresses the rank inconsistency problem found in methods such as OR-NN. Similarly to methods such as OR-CNN, the ordinal regression problem is transformed into $K - 1$ binary classification tasks, where $f_k(x_i) \in \{0, 1\}$ denotes the output of the $k$-th binary classifier. The final predicted rank is obtained by summing up the outputs of the binary classifiers:

\begin{equation}
    rank = 1 + \sum_{k=1}^{K-1} f_k(x_i)
\end{equation}

To ensure consistent predictions, CORAL enforces rank monotonicity, that is, the outputs must satisfy: $f_1(x_i) \geq f_2(x_i) \geq \dots \geq f_{K-1} (x_i)$.
Consistency is achieved by sharing weights parameters across all $K-1$ binary tasks but with independent bias.

Let $W$ denote the shared weights of the penultimate layer of the neural network, and let $g(x_i, W)$ represent its output. The final layer adds independent biases $b_k$ that produce logits $\{ g(x_i, W) + b_k \}$. All $K-1$ outputs are passed through a sigmoid activation to compute the probabilities. The model is trained using the cross-entropy loss over the $K - 1$ binary classifiers:

\begin{equation}
    L(W, b) = -\sum_{i=1}^{N} \sum_{k = 1}^{K - 1} \lambda^{(k)} [log(\sigma( g({x_i},  W) + b_k )) y_i^{(k)} + log(1 - \sigma( g({x_i},  W) + b_k )  (1 - y_i^{(k)})]
\label{equation:coral-loss}
\end{equation}

In most practical scenarios, task weights are uniform $\lambda^{(k)} = 1$. The authors prove that minimizing this loss function results in non-increasing biases: $b_i \geq b_2 \geq \dots \geq b_{K-1}$. This ensures that the classifier outputs are also non-increasing, leading to rank-monotonic and consistent predictions.

\subsubsection{Conditional Ordinal Regression for Neural networks (CORN)}

CORN \cite{shi2021deep} guarantees rank consistency without relying on the weight-sharing constraint used in CORAL. Although the shared weights constraint in CORAL ensures consistency, it can reduce the expressiveness of the model. CORN avoids this limitation by using a training scheme based on conditional training subsets and the chain rule of probability.

As in CORAL, the ordinal regression problem is decomposed into $K - 1$ binary tasks. However, CORN estimates a series of conditional probabilities using nested events: $\{ y_i > rank_k\} \subseteq \{ y_i > rank_{k-1} \}$. The output of the $k$-th classifier $f_k(x_i)$ is defined as:

 \begin{equation}
     f_k(x_i) =
     \begin{cases}
         P(y_i > rank_k | y_i>rank_{k-1}) & \text{if } k > 1 \\
        P(y_i> rank_i) & k =1
     \end{cases}
 \end{equation}

The unconditional probabilities can be reconstructed using the chain rule:

 \begin{equation}
     P(y_i>rank_k) = \prod_{j = 1}^k f_j(x_i)
    \label{equation:corn-prob}
 \end{equation}

Because $\forall j \text{, } 0 \leq f_j(x_i) \leq 1$, the resulting probabilities are guaranteed to be non-increasing, which ensures rank consistency in the final outputs.

 \begin{equation}
     P(y_i > rank_1) \geq P(y_i \geq rank_2) \geq \dots \geq P(y_i > rank_{K-1})
 \end{equation}

To train conditional probabilities, CORN constructs training subsets specific to each binary task. Each subset $S_k$ is used to train the prediction of $P(y_i > rank_k | y_i > rank_{k - 1})$.

 \begin{equation}
     S_k = \{(x_i, y_i) | y_i > rank_{k-1}\} \text{ for  } k \in \{1, \dots, K - 1\}
 \end{equation}
 
 The loss function minimized during backpropagation is:

 \begin{equation}
     L(X, y) = - \frac{\sum_{j = 1}^{K-1} \sum_{i = 1}^{|S_j|} [log(f_j(x_i)) * \mathbb{1} \{y_i > rank_j\} + log(1 - f_j(x_i)) * \mathbb{1} \{y_i \leq rank_j\}]}{\sum_{j =1}^{K - 1}|S_j|}
 \end{equation}

To improve numerical stability, the authors also introduce an equivalent formulation in terms of logits $z_i$, where $log(\sigma(z_i)) = log(f_j(x_i))$:

\begin{equation}
     L(X, y) = - \frac{\sum_{j = 1}^{K-1} \sum_{i = 1}^{|S_j|} [log(\sigma(z_i)) * \mathbb{1} \{y_i > rank_j\} + ( log(\sigma(z_i)) - z_i) * \mathbb{1} \{y_i \leq rank_j\}]}{\sum_{j =1}^{K - 1}|S_j|}
 \end{equation}

Similarly to CORAL, the final predicted rank is computed by counting the number of outputs greater than 0.5, using the same formula.

\subsubsection{CONDitionals for Ordinal Regression (CONDOR)}

CONDOR (CONDitionals for Ordinal Regression) \cite{jenkinson2021universally}
 approach reformulates the ordinal regression problem as a series of binary classification tasks. Unlike methods such as Simple Ordinal Regression, which rely on multiple independent models, CONDOR leverages a specific architecture and loss for neural network, which guarantee model outputs consistent with ordinal nature of the problem.
 This method was developed to address key limitations of the CORAL framework, particularly its reliance on shared weights. Importantly, CONDOR ensures rank-monotonic probability outputs. The final class prediction is computed identically to CORAL and CORN, by summing the binary outputs across thresholds.

As in other ordinal approaches, CONDOR transforms the ordinal regression task into $K - 1$ binary classification problems: $f_k:X \rightarrow \{0, 1\}$. Each target class is encoded as:

 \begin{equation}
\text{class(y)} = [1 \text{ if } i \leq y \text{ else } 0] \quad \text{for } i \in \{1, \dots, K-1\}
\end{equation}

Binary predictions are obtained by thresholding the predicted probabilities: $f_k(x) = \mathbb{1}_{p_k(x) > 0.5} $ where $p_k: \mathbb{X} \rightarrow [0, 1]$ represents the predicted probability. In the case of neural networks, these functions are parameterized with their weights $\theta \in \Theta$, resulting in the model (function) $f_k(x;\theta)$, predicting probabilities $p_k(x;\theta)$.

Rather than estimating marginal probabilities, CONDOR estimates conditional probabilities, similar to CORAL:

\begin{equation}
    q_k(x, \theta) = P(y_k = 1| \text{x} = x, y_{k-1} = 1, \theta = \theta)
\end{equation}

The boundary condition $y_0 = 1$ is set to unit probability. The marginal probabilities are then obtained using the product rule, such as for CORN (see Equation \ref{equation:corn-prob}):

\begin{equation}
    p_k(x, \theta) = \prod_{k'=1}^k q_{k'}(x, \theta)
\end{equation}

These equations can be applied to any model that produces binary probabilities. In CONDOR, each output node represents one of the $q_k(x, \theta)$ values and uses a sigmoid activation to ensure the output lies in $(0, 1)$.

Rank consistency is enforced by construction. Since each $q_k(x, \theta)$ is in range $(0, 1)$, the resulting marginal probabilities $p_k(x; \theta)$ are guaranteed not to increase.

For model training, CONDOR uses a loss function derived from maximum likelihood estimation. Let $y_k^{(i)}$ be the binary label for the $k$-th task. The loss function is:

\begin{equation}
    L(\theta) = -\sum_{i=1}^{N} \sum_{k = 1}^{K - 1} y_{k-1}^{(i)}(log(q_k(x_i;\theta)) y_{k}^{(i)} + log(1 - q_k(x_i;\theta)) (1 - y_{k}^{(i)}))
\end{equation}

This formulation ensures that only the conditional probabilities $q_k$ associated with positive $y_{k - 1}^{(i)} = 1$ contribute to the loss. The authors prove that minimizing this loss yields the maximum likelihood estimate of the parameters $\theta$.
A key advantage of CONDOR is its universality: it guarantees rank-consistent output regardless of the upstream neural network architecture, provided the final layer adheres to the CONDOR formulation. This makes it broadly applicable and theoretically robust compared to methods like CORAL, which require specific weight-sharing constraints.

\section{Evaluation methodology}

The dataset used in this work is an example of tabular data. Although it is not time series data, the records, i.e., monsters, were published sequentially over time as part of official rulebooks. Therefore, it is important to consider the chronological nature of the dataset during evaluation to ensure a realistic evaluation of model performance. The chosen evaluation strategy must capture both the characteristics of tabular ordinal regression and the chronological structure of the data. This is essential for accurately estimating the generalization capabilities of a model and for preventing data leakage between training and test sets.

To satisfy these requirements, a chronological data split was used instead of the classic random hold-out method. To further assess generalization, an expanding window evaluation, inspired by time series validation techniques, was also implemented. In addition, appropriate metrics for ordinal regression were applied to evaluate how well the predicted and true label orderings align. Given that ordinal regression lies at the intersection of classification and regression, evaluation metrics from both problems were used. These include metrics that capture the number of exact predictions and the distance between the predicted and actual ordinal classes. For hyperparameter tuning, a standard cross-validation (CV) procedure was used.

\subsection{Train-test split evaluation}

The basic evaluation method uses a single train-test split, also known as a hold-out method. Two types of data splitting strategies were tested: random and chronological.

Random splitting assigns data to sets at random. This method is widely used because of its simplicity and efficiency. However, it does not account for the structure or temporal nature of the data and does not capture the fact that monster designers often base new creations on existing ones. As a result, random splitting can introduce data leakage, where information from similar or related instances appears in both training and test sets. This leads to an overly optimistic estimate of the generalization performance of the model, which undermines the reliability of the evaluation.

Chronological splitting, on the other hand, assigns older data to the training set and the latest data to the test set. This approach reflects the natural emergence of new data over time. It is more realistic because it simulates predicting future values using a model trained on past data. 

In Pathfinder, monsters are introduced in batches within consecutively published rulebooks, so the dataset exhibits a natural chronological order. In this case, the training set consists of monsters from previous publications, while the test set includes the most recent monsters from the publisher’s official database.

\subsection{Expanding window evaluation}

Due to the chronological nature of the dataset, an additional evaluation method was introduced: the expanding window approach. This method is inspired by standard evaluation strategies used in time series forecasting. The dataset was partitioned into chronologically ordered subsets. The first subset included all monsters published at the premiere of Pathfinder's second edition. Each subsequent subset included at least $N = 100$ monsters that were not present in any of the previous subsets, selected according to the oldest remaining publication dates. This emulates the design phase for one or more new books with monsters. To ensure temporal consistency, all monsters published on the same day were always assigned to the same subset. This procedure produced $W$ subsets, each corresponding to a chronological ``window'' of data. The models were evaluated using an expanding window strategy: for each $w \in \{1, \dots, W - 1\}$, the model was trained in the union of subsets $S_1 \bigcup S_1 \bigcup \dots \bigcup S_w$ and tested in the subset $w+1$. 
The final results of the evaluation are reported as the mean and standard deviation of the performance of the model in these $W - 1$ iterations. This setup provides a more realistic estimate of generalization performance over time, reflecting how models are expected to predict new unseen monsters based on previously published ones. Due to multiple test sets, this setting allows the computation of standard deviation and a more stable estimation of generalization performance, compared to a holdout strategy.

\subsection{Metrics}

The performance of the models was evaluated using several metrics: RMSE, MAE, accuracy, accuracy at $k$, and Somers' D. While RMSE and MAE are standard metrics for classical regression, they are also suitable for ordinal regression tasks \cite{gutierrez2012experimental, saito2021evaluation, hu2018collaborative}, as they account for the ordered nature of the output classes and penalize errors based on their magnitude. Detailed descriptions of those metrics are provided in the following subsections.

The dataset used is imbalanced, with a significantly higher number of monsters in lower levels compared to higher levels. As a result, the MAE and RMSE values may not fully reflect the quality of the model predictions. To address this, macro-averaged versions of these metrics were also employed \cite{baccianella2009evaluation}.

Accuracy is a very harsh metric, because it only counts predictions as correct when the predicted class matches the true class exactly, without considering how close the incorrect predictions are. However, in case of an ordinal regression problem, the magnitude of an error depends on the distance between the true and predicted classes. To address this limitation, accuracy@k was also used, which considers a prediction correct if it falls within the distance of k classes to the true label. This approach is similar to the top-k accuracy commonly used in computer vision tasks, where predictions are considered correct if the true label is among the top-k predicted classes.

Additionally, Somers' D was used to assess the ordinal association between predictions and ground-truth labels. It is a measure specifically designed for ordinal dependent variables and, for this reason, is commonly used in the case of an ordinal regression problem \cite{o2006logistic, harrell2015ordinal}.

\subsubsection{Root Mean Squared Error (RMSE)}

RMSE gives more weight to larger errors due to the squared term, making it particularly sensitive to significant prediction mistakes. The square root ensures that the result is in the same unit as the original data. It is defined as:

\begin{equation}
\text{RMSE} = \sqrt{\frac{1}{n} \sum_{i=1}^{n} \left(y_i - \hat{y}_i \right)^2}
\end{equation}

The macro-averaged version computes RMSE \cite{baccianella2009evaluation} for each class separately and then averages them:

\begin{equation}
\text{RMSE}^M = \frac{1}{k} \sum_{j=1}^{k} \sqrt{ \frac{1}{N_j} \sum_{y_i = j} \left(y_i - \hat{y}_i \right)^2 }
\end{equation}

Macro-averaged metrics ensure that all classes contribute equally, which is especially useful for imbalanced datasets. For perfectly balanced data, unweighted and macro-averaged RMSE are the same.

\subsubsection{Mean Absolute Error (MAE)}

MAE treats all errors in direct proportion to their magnitude, making it more interpretable and less sensitive to outliers than RMSE:

\begin{equation}
\text{MAE} = \frac{1}{n} \sum_{i=1}^{n} \left| y_i - \hat{y}_i \right|
\end{equation}

The macro-averaged MAE is computed analogously to the macro-averaged RMSE:

\begin{equation}
\text{MAE}^M = \frac{1}{k} \sum_{j=1}^{k} \frac{1}{N_j} \sum_{y_i = j} \left| y_i - \hat{y}_i \right|
\end{equation}

\subsubsection{Accuracy}

Accuracy measures the proportion of exactly correct predictions:

\begin{equation}
\text{Accuracy} = \frac{1}{n} \sum_{i=1}^{n} \mathds{1}_{\{y_i = \hat{y}_i\}}
\end{equation}

Although it is a common classification metric, it is not ideal for ordinal regression, since it ignores the magnitude of error. As such, it may penalize near-correct predictions too harshly. To give a more forgiving view of ordinal predictions, accuracy@k \cite{gaudette2009evaluation} considers a prediction correct if it falls within $k$ levels of the true label:

\begin{equation}
\text{Accuracy@k} = \frac{1}{n} \sum_{i=1}^{n} {1}(|y_i - \hat{y}_i| \leq k)
\end{equation}

Based on this definition, Accuracy@0 corresponds to standard accuracy. Accuracy@1 allows for a one-level deviation (either up or down), which is often a more informative and practical metric in ordinal settings. Furthermore, the difference between accuracy@1 and accuracy@0 quantifies how many predictions were close but not exact, offering insight into close predictions rather than treating them as total failures.

\subsubsection{Somers' D}

Somers' D \cite{somers1962new} is a rank correlation metric designed for ordinal variables. It evaluates the agreement between the ordering of the predicted and true labels. The value ranges from $-1$ (perfect disagreement) to $1$ (perfect agreement). Two pairs $(x_i, y_i)$ and $(x_j, y_j)$ are:

\begin{itemize}
    \item Concordant if $(x_i > x_j \text{ and } y_i > y_j)$ or $(x_i < x_j \text{ and } y_i < y_j)$
    \item Discordant if $(x_i > x_j \text{ and } y_i < y_j)$ or $(x_i < x_j \text{ and } y_i > y_j)$
    \item Tied if $x_i = x_j$ or $y_i = y_j$
\end{itemize}

In the context of ordinal regression, the pairs $(x_i, y_i)$ and $(x_j, y_j)$ represent predictions and true labels for two different instances $i$ and $j$. Each pair $(x, y)$ is a (prediction, label).

Somers' D is defined as:

\begin{equation}
\text{Somers'}D = \frac{N_C - N_D}{N_C + N_D + N_T}
\end{equation}

Where:
\begin{align*}
N_C &= \text{number of concordant pairs} \\
N_D &= \text{number of discordant pairs} \\
N_T &= \text{number of tied pairs}
\end{align*}

Alternatively, it can be expressed using the Kendall tau rank correlation coefficient:

\begin{equation}
\tau_A = \frac{n_c - n_d}{n_0}, \quad \text{where } n_0 = \frac{n(n - 1)}{2}
\end{equation}

\begin{equation}
\text{Somers'}D = \frac{\tau_A(X, Y)}{\tau_A(Y, Y)}
\end{equation}

\chapter{Experiments and results}

This chapter presents the experimental setup, including the feature engineering process, model training procedures, evaluation methodology, and the results obtained from the experiments. 

\section{Experimental setup}

This section outlines the configuration of the experimental environment and details the process of constructing and extracting the dataset, training models, and evaluation strategies.

\subsection{Dataset}

As part of this work, a dataset that contains Pathfinder Second Edition monsters was constructed. A detailed explanation of the feature engineering process is provided in Section \ref{section:methods_dataset}. The final dataset comprises 2,637 unique monster entries, each described with 32 features that capture a wide range of characteristics that describe each monster. These features include:
\begin{itemize}
    \item Core attributes: Strength, Dexterity, Constitution, Intelligence, Wisdom, and Charisma;
    \item Armor Class (AC) — a measure of how effectively the monster can avoid attacks;
    \item Hit Points (HP) — indicating the amount of damage a monster can withstand;
    \item Saving throws: Will, Fortitude, and Reflex — representing resistances to various effects;
    \item Perception — the monster’s ability to detect threats;
    \item Movement speeds: land, flying, and swimming speeds;
    \item Melee and ranged attack capabilities, including maximum attack bonus and average damage;
    \item Number of immunities;
    \item Spellcasting-related features: number of spells per level, spell difficulty class (DC), and spell attack bonus.
\end{itemize}

In addition to the feature set, the dataset includes two key target variables: the monster's level (used as the dependent variable for ordinal regression tasks) and the source rulebook in which the monster was originally published. The latter is used to implement a chronological data split, preserving the temporal structure of the dataset for evaluation purposes.
As described in Section \ref{section:methods_dataset}, the predicted level of a monster ranges from -1 to 21.

\subsection{Model training and evaluation}

\subsubsection{Train-test split and tuning}

The dataset used in this study is chronological in nature; see Section \ref{section:methods_dataset}. Therefore, the evaluation methodology was designed to respect both the tabular format and the temporal structure of the data. Two evaluation strategies were used: 
\begin{itemize}
    \item Chronological train-test split evaluation - the training set consists of monsters appearing in earlier sourcebooks, while the test set includes only monsters from more recent publication periods.
    \item Expanding window evaluation - time series inspired evaluation strategy. At each step, the training set includes all the preceding subsets, and the test set consists of the next chronological window. 
\end{itemize}

All models were tuned using cross-validation in conjunction with a grid search over hyperparameters specific to each model. For the final analysis, a subset of all the evaluated models was selected. This subset includes the most representative models from groups of similar approaches or those that yielded particularly interesting results.

\subsubsection{Metrics}

The task of ordinal regression combines aspects of both classification and classical regression. Consequently, a combination of evaluation metrics from both domains was used to comprehensively assess the performance of the model.
To capture the magnitude of prediction errors, traditional regression-based metrics, including RMSE and MAE, were used. Both were reported as the macroaveraged version (denoted $MAE^M$ and $RMSE^M$) due to the class imbalance present in the Pathfinder dataset. Classification-style metrics were applied to assess the precision of label predictions, to measure exact match (accuracy), and a variation that allows an error margin of one class above or below the true label (accuracy@1). To further assess the ordinal relationship between the predicted and actual labels, the Somers’ D rank correlation coefficient was used. This metric evaluates the agreement between the ordering of predicted values and true labels.

\subsubsection{Models}

A baseline model was developed based on domain-specific knowledge and the guidelines provided by the Pathfinder Second Edition authors for constructing monsters. This model emulates the human approach to estimating the level of monsters. Specifically, it assigns each monster an archetype (i.e., a general monster type) based on the distribution of its statistics. Then a $k$-nearest neighbors (KNN) model is trained separately for each archetype and used to predict the level of the monster within that group.

The first strategy to address the ordinal regression task involved using classical regression models to predict continuous values that were later mapped to discrete levels. This two-step approach consists of a regression model followed by a rounding strategy. The set of models evaluated under this framework includes the following:
\begin{itemize}
    \item Linear models (with the best results for LASSO regression);
    \item Support Vector Machines (SVM) with a radial basis function (RBF) kernel;
    \item Random Forests (RF);
    \item Light Gradient Boosting Machine (LightGBM).
\end{itemize}
To map the continuous outputs of these models to ordinal levels, four rounding strategies were employed:
\begin{itemize}
    \item Standard mathematical rounding, with a fixed threshold of 0.5;
    \item Tuned global rounding, where a single threshold value is optimized across all predictions;
    \item Level-specific thresholds optimized via:
    \begin{itemize}
        \item Tree-structured Parzen Estimator (TPE), implemented with Optuna, treating each threshold as a hyperparameter;
        \item A mathematical programming formulation based on the shortest path in a graph, designed to find optimal thresholds between neighbouring levels.
    \end{itemize}
\end{itemize}
Two discrete threshold sets were evaluated in the tuning process: $R_1= [0.05, 0.1, 0.15, \dots, 0.95]$ and $R_2=  [0.25, 0.3, 0.35, \dots, 0.75]$.
Since the TPE method operates over a continuous space, only the minimum and maximum values from these sets were used to define its search boundaries. The thresholds identified through these optimization techniques were subsequently used to map the prediction of the model to ordered labels.  

The second modeling strategy involves the use of algorithms specifically designed for ordinal regression tasks. These models inherently incorporate the ordinal nature of the target variable and directly predict discrete ordered labels. The detailed description of those models is given in Section \ref{sec:ordinal_regression_methods}. The set of ordinal models evaluated:
\begin{itemize}
    \item Simple Approach to Ordinal Classification (SAOC),
    \item Ordered Logistic Regression (ORD),
    \item Threshold-based model - Immediate-Threshold (LogisticIT),
    \item Gaussian Processes for Ordinal Regression (GPOR),
    \item Ordinal Random Forests (ORF),
    \item Deep Neural Networks for Ordinal Regression
    \begin{itemize}
        \item CORAL,
        \item CORN,
        \item Spacecutter,
        \item NNRank,
        \item CONDOR,
        \item OR-CNN
    \end{itemize}
\end{itemize}

\section{Results and discussion}

This section presents and analyzes the results of the models evaluated using different performance metrics for each split type. Shows results and compares them on different levels/for different aspects for each evaluation method separately.
For clarity and conciseness, only a single rounding variant per model is presented in the figures below. Specifically, the variant that achieved the best result for a given primary metric (macroaveraged MAE) was selected for visualization. This ensures a fair comparison of each model while avoiding unnecessary redundancy. It is important to note that while different rounding strategies may yield slightly better results for specific metrics, the results of the chosen rounding strategy were reported for all metrics. 

In all result tables, the top three performing models for each metric are highlighted using a color-coded scheme for ease of interpretation:
\begin{enumerate}
    \item Green - best result
    \item Blue - second-best
    \item Violet - third-best
\end{enumerate}

\subsection{Train-test split}

\subsubsection{Overall models performance comparison}

\begin{table}[H]
  \centering
\begin{tabular}{|l||>{\centering\arraybackslash}m{2cm}|>{\centering\arraybackslash}m{2cm}|c|c|c|}
  \hline
    \textbf{Model} & \textbf{$\mathbf{MAE}^{\mathbf{M}}$} & \textbf{$\mathbf{RMSE}^{\mathbf{M}}$} & \textbf{Somer's D} & \textbf{Acc} & \textbf{Acc@1} \\
  \hline \hline
      \cellcolor{red!25}Baseline & 0.82 & 1.37 & 0.91 & 46\% & 86\% \\
      LR & 0.42 & 0.80 & \cellcolor{blue!25}0.96 & 67\% & \cellcolor{blue!25}97\%\\
      SVM  & 0.35 & 0.72 & \cellcolor{green!50}0.97 & \cellcolor{violet!25}73\% & \cellcolor{green!50}98\% \\
      \cellcolor{blue!10}RF & \cellcolor{violet!25} 0.27 & \cellcolor{green!50}0.59 &  \cellcolor{green!50}0.97 & \cellcolor{blue!25}77\% & \cellcolor{green!50}98\% \\
      \cellcolor{blue!10}LightGBM & 0.28 & \cellcolor{blue!25}0.62 & \cellcolor{green!50}0.97 & \cellcolor{blue!25}77\% & \cellcolor{green!50}98\% \\
      \hline
      ORD & 0.36 & 0.74 & \cellcolor{green!50}0.97 & 70\% & \cellcolor{blue!25}97\% \\
      \cellcolor{blue!10}ORF & \cellcolor{green!50} 0.24 & \cellcolor{blue!25} 0.61 & \cellcolor{green!50}0.97 & \cellcolor{green!50}81\% & \cellcolor{blue!25}97\% \\
      LogisticIT & 0.37 & 0.73 & \cellcolor{green!50}0.97 & 70\% & \cellcolor{blue!25}97\% \\
      \cellcolor{green!25}SAOC & \cellcolor{blue!25} 0.25 & \cellcolor{green!50} 0.59 & \cellcolor{green!50}0.97 & \cellcolor{green!50}81\% & \cellcolor{blue!25}97\% \\
      GPOR & 0.34 & 0.70 & \cellcolor{green!50}0.97 & 72\% & \cellcolor{blue!25}97\% \\
      CORAL & 0.41 & 0.76 & \cellcolor{blue!25}0.96 & 68\% & \cellcolor{blue!25}97\% \\
      CORN & 0.42 & 0.79 & \cellcolor{blue!25}0.96 & 65\% & \cellcolor{blue!25}97\% \\
      Spacecutter & 0.59 & 0.91 & \cellcolor{violet!25}0.95 & 51\% & 93\% \\
      NNRank & 0.58 & 0.87 & \cellcolor{blue!25}0.96 & 50\% & \cellcolor{violet!25}95\% \\
      CONDOR & 0.46 & 0.82 & \cellcolor{blue!25}0.96 & 64\% & \cellcolor{violet!25}95\% \\
      OR-CNN & 0.42 & 0.79 & \cellcolor{green!50}0.97 & 67\% & \cellcolor{blue!25}97\% \\
      \hline
\end{tabular}
      \caption{Results of all models.}
      \label{tab:results_models_all}
\end{table}

\begin{table}[h]
  \centering
\begin{tabular}{|l||>{\centering\arraybackslash}m{2cm}|>{\centering\arraybackslash}m{2cm}|c|c|c|}
  \hline
    \textbf{Model} & \textbf{$\mathbf{MAE}^{\mathbf{M}}$} & \textbf{$\mathbf{RMSE}^{\mathbf{M}}$} & \textbf{Somer's D} & \textbf{Acc} & \textbf{Acc@1} \\
  \hline \hline
      ORF & 0.24 & 0.61 & 0.97 &81\% & 97\% \\
      SAOC &0.25 & 0.59 &  0.97 & 81\% & 97\% \\
      RF & 0.27 & 0.59 &0.97 & 77\% &98\% \\
      LightGBM & 0.28 & 0.62 & 0.97 & 77\% & 98\% \\
      GPOR & 0.34 & 0.7 & 0.97 & 72\% & 97\% \\
      \hline
\end{tabular}
      \caption{Top-5 models, according to macroaveraged MAE.}
      \label{tab:results_models_best}
\end{table}

First, we compare the overall results for all models; see Table \ref{tab:results_models_all}. 
Among all models evaluated, the baseline approach yielded significantly inferior results compared to more complex ML methods. This simple human-inspired strategy, which relies on predetermined archetypes and KNN predictions, struggles to accurately estimate monster levels. A key issue is that monsters sharing the same archetype, and even appearing as immediate neighbors in the dataset, may exhibit substantially different levels due to nuanced differences in their abilities and statistics. As a result, the baseline model performed poorly in all evaluation metrics. Even in the case of accuracy@k, where most ML models achieved results exceeding 90\%, the baseline lagged by more than 10 percentage points in most cases.

In contrast, the best-performing models across all metrics were consistently tree-based methods. These included Random Forest (RF), LightGBM, and specialized ordinal variants such as Ordinal Random Forest (ORF) and SOR Random Forest (SAOC). For different rounding strategies applied, these models dominated the top ranks in terms of macroaveraged RMSE, MAE, Somer's D, and classification accuracy. The overall best-performing model was the ordinal regression forest, which uses a random forest as its base learner within an ordinal regression framework. It achieved the best results in macroaveraged RMSE, Somers’ D, and accuracy. The only metric in which it did not rank as one of the best three was accuracy@1, where its score was only 1 percentage point lower than the top model, still an impressive performance around 97\%. However, overall all tree-based models got really similar results.

When comparing dedicated ordinal regression models to classical models with rounding, no clear winner emerges. Both groups contain models that perform exceptionally well and those that fall near the bottom of the ranking. However, there is a notable subgroup of ordinal regression models based on neural networks that consistently underperform relative to other approaches. Models such as Spacecutter and NNRank, which do not explicitly enforce ranking consistency or incorporate ordinal weighting during training or prediction, yielded the weakest results among all non-baseline models.

Interestingly, more advanced neural ordinal models such as CORAL, CONDOR failed to outperform simpler ordinal classification techniques such as the Simple Approach to Ordinal Classification (SAOC) when used with well-established classifiers like RF. Although they were designed to mitigate issues like rank inconsistency. However, they achieve better performance than the simplest neural models such as NNRank. At the same time, the OR-CNN model, which is basically the same as NNRank but with an additional weighting factor added to the loss function, outperforms those more sophisticated methods.

\subsubsection{Accuracy}

\begin{table}[h]
  \centering
\begin{tabular}{|l||>{\centering\arraybackslash}m{2cm}|>{\centering\arraybackslash}m{2cm}|c|c|c|}
  \hline
    \textbf{Model} & \textbf{Acc} & \textbf{Acc@1} \\
  \hline \hline
      Baseline & 46\% & 86\% \\
      LR & 67\% & 97\%\\
      SVM  & 73\% &98\% \\
      RF & 77\% & 98\% \\
      LightGBM & 77\% & 98\% \\
      \hline
      ORD & 70\% & 97\% \\
      ORF & 81\% & 97\% \\
      LogisticIT & 70\% & 97\% \\
      SAOC & 81\% & 97\% \\
      GPOR & 72\% & 97\% \\
      CORAL & 68\% & 97\% \\
      CORN & 65\% & 97\% \\
      Spacecutter & 51\% & 93\% \\
      NNRank & 50\% & 95\% \\
      CONDOR & 64\% & 95\% \\
      OR-CNN & 67\% & 97\% \\
      \hline
\end{tabular}
      \caption{Accuracy results.}
      \label{tab:results_acc}
\end{table}

The general results of the accuracy@1 metric, see Table \ref{tab:results_acc}, compared to standard accuracy, suggest that even when a model fails to predict the exact level of a monster, it often predicts a level that is only one class away from the correct one. For nearly all models, accuracy@1 exceeds 95\%. This means that models rarely misclassify by more than one level. In contrast, the standard accuracy exhibits much greater variability, ranging from approximately 50\% to 81\%, depending on the model and the rounding strategy used. It should be noted that both ordered models from this group (SAOC and ORF) have an accuracy greater than 80\%, while other tree models are 4 p.p. below it. Non-tree models get substantially lower accuracy, around 10 p.p. lower than the best tree-based ones. 

The relatively small variance in accuracy@1 compared to the larger spread in standard accuracy is particularly noteworthy. It suggests that while most models are effective at producing predictions close to the correct label, only the best models make precise predictions with high frequency. This observation may imply that achieving high accuracy@1 is a necessary, though not sufficient, condition for strong overall performance in this specific task.

\subsubsection{Rounding strategies comparison}
\label{sec:results_rounding}

In this section, the color scheme is applied row-wise rather than column-wise.

\begin{table}[H]
  \centering
\begin{tabular}{|l||c|c|c|c|c|c|c|}
  \hline
    {\textbf{Model}} &
     {\makecell{\textbf{Round}\\\textbf{0.5}}} &
     {\makecell{\textbf{Global}\\\textbf{$R_1$}}} &
     {\makecell{\textbf{TPE}\\\textbf{$R_1$}}} &
     {\makecell{\textbf{Graph}\\\textbf{$R_1$}}} &
     {\makecell{\textbf{Global}\\\textbf{$R_2$}}} &
     {\makecell{\textbf{TPE}\\\textbf{$R_2$}}} &
    {\makecell{\textbf{Graph}\\\textbf{$R_2$}}} \\
  
  \hline
  \hline
    Baseline & 0.861 & 0.861 & 0.853 & \cellcolor{green!50}0.823 & 0.861 & 0.864 & \cellcolor{green!50}0.823  \\
    
    LR & 0.425 & 0.425 & 0.433 & 0.427 & 0.425 & 0.428 & \cellcolor{green!50}0.424 \\

    SVM & 0.354 & 0.361 & 0.365 & \cellcolor{green!50}0.350 & 0.361 & 0.356 & 0.363 \\
    
    RF & 0.277 & 0.295 & 0.298 & \cellcolor{green!50}0.274 & 0.295 & 0.289 & 0.276 \\
    
    LightGBM & \cellcolor{green!50}0.282 & 0.492 & 0.366 & 0.483 & 0.329 & 0.295 & 0.324 \\
    
    GPOR & 0.381 & 0.381 & 0.398 & \cellcolor{green!50}0.343 & 0.381 & 0.386 & 0.346 \\
      \hline
\end{tabular}
      \caption{Results of rounding strategies.}
      \label{tab:results_rounding}
\end{table}

Different rounding strategies yielded similar results for a given model. In typical cases, variations between rounding methods amount to no more than 2 percentage points (p.p.) in accuracy and approximately 0.02 in macroaveraged RMSE or MAE. However, notable exceptions were observed, particularly with the LightGBM model.

For LightGBM, the differences between rounding strategies were significantly more pronounced, with variations of up to 20 p.p. in accuracy and 0.2 in macroaveraged MAE and RMSE. This deviation is attributed to LightGBM's tendency to overfit, producing predictions that are almost exact integers on the training set. Since threshold optimization procedures are performed using training predictions, this overfitting leads to suboptimal threshold selection for global threshold tuning and the graph-based method. For those strategies, thresholds are often driven to extremely low values, and as a consequence compound final results. As a result, for LightGBM, the most reliable and effective rounding strategy was simple mathematical rounding using a fixed threshold of 0.5. Among the threshold-tuning methods, the strategy that treats thresholds as hyperparameters (TPE) noticeably outperformed both the global threshold-tuning and the graph-based approach. This is probably also due to the nature of the LightGBM overfitting: a wide range of thresholds between two adjacent levels can produce identical results in the training set. The TPE method, being stochastic in nature, is more likely to identify near-optimal thresholds by chance.

For the remaining models, the best results were generally achieved using the graph-based threshold tuning strategy. Specifically, this approach performed best using the threshold group $R_1$ for SVM, RF, and GPOR, $R_2$ for LR (LASSO) and either of them for the baseline. 

It is worth noting that the tuned global threshold strategy yielded the worst results, not only for LightGBM, but also for all other models (excluding Baseline), including high underperformance compared to simple mathematical rounding.  Furthermore, for the accuracy@1 metric, the differences between the various rounding strategies were minimal - typically within tenths of a percentage point - indicating the relative robustness of this metric to the choice of rounding and showing that most distances from incorrectly predicted values are minimal.

In conclusion, there is no universally superior rounding strategy. The optimal choice depends on the specific characteristics of the model being used. For models prone to overfitting, such as LightGBM, mathematical rounding with a 0.5 threshold is the most stable and reliable option. However, for models that are more robust to this, tuning thresholds per level, particularly using the graph-based approach, can yield the best performance. However, it should also be noted that this tuning thresholds per level strategy (when implemented via TPE) can lead to worse results than mathematical rounding. Thus, the selection of the rounding strategy should depend both on the behavior of the model and on the nature of the threshold optimization method.

\subsubsection{Comparison of Regular and Macroaveraged}
\label{sec:regular_vs_macroaveraged}

\begin{table}[H]
    \centering
\begin{tabular}{|l||>{\centering\arraybackslash}m{2cm}|>{\centering\arraybackslash}m{2cm}|c|c|c|}
  \hline
  \textbf{Model} & \textbf{MAE} & \textbf{$\mathbf{MAE}^{\mathbf{M}}$} & \textbf{RMSE} & \textbf{$\mathbf{RMSE}^{\mathbf{M}}$} \\
  \hline
  \hline
    Baseline & 0.793 & 0.861 & 1.312 & 1.409 \\
    
    LR & 0.367 & 0.425 & 0.729 & 0.799 \\
    
    SVM & 0.301 & 0.354 & 0.627 & 0.707 \\
    
    RF & 0.236 & 0.277 & 0.559 & 0.603 \\
    
    LightGBM & 0.260 & 0.282 & 0.603 & 0.616 \\

    \hline
    
    ORD & 0.342 & 0.368 & 0.697 & 0.742 \\
    
    ORF & 0.230 & 0.241 & 0.616 & 0.609 \\
    
    LogisticIT & 0.344 & 0.369 & 0.694 & 0.733 \\
    
    SAOC & 0.224 & 0.245 & 0.572 & 0.588 \\
    
    GPOR & 0.321 & 0.343 & 0.659 & 0.697 \\
    
    CORAL & 0.361 & 0.405 & 0.699 & 0.757 \\
    
    CORN & 0.394 & 0.422 & 0.735 & 0.786 \\
    
    Spacecttter & 0.569 & 0.594 & 0.876 & 0.912 \\
    
    NNRank & 0.563 & 0.576 & 0.853 & 0.874 \\
    
    CONDOR & 0.416 & 0.457 & 0.762 & 0.821 \\
    
    OR-CNN & 0.374 & 0.424 & 0.721 & 0.787 \\
    \hline
\end{tabular}
    \caption{Regular and macroaveraged MAE and RMSE values.}
    \label{tab:results_rmse_mae}
\end{table}

As expected, the results for macro-averaged variants of MAE and RMSE are generally worse (higher) than their regular counterparts. However, the magnitude of the difference varies between models. This behavior is expected given the class imbalance.

\begin{figure}[htp]
    \centering
    \includegraphics[width=0.75\textwidth]{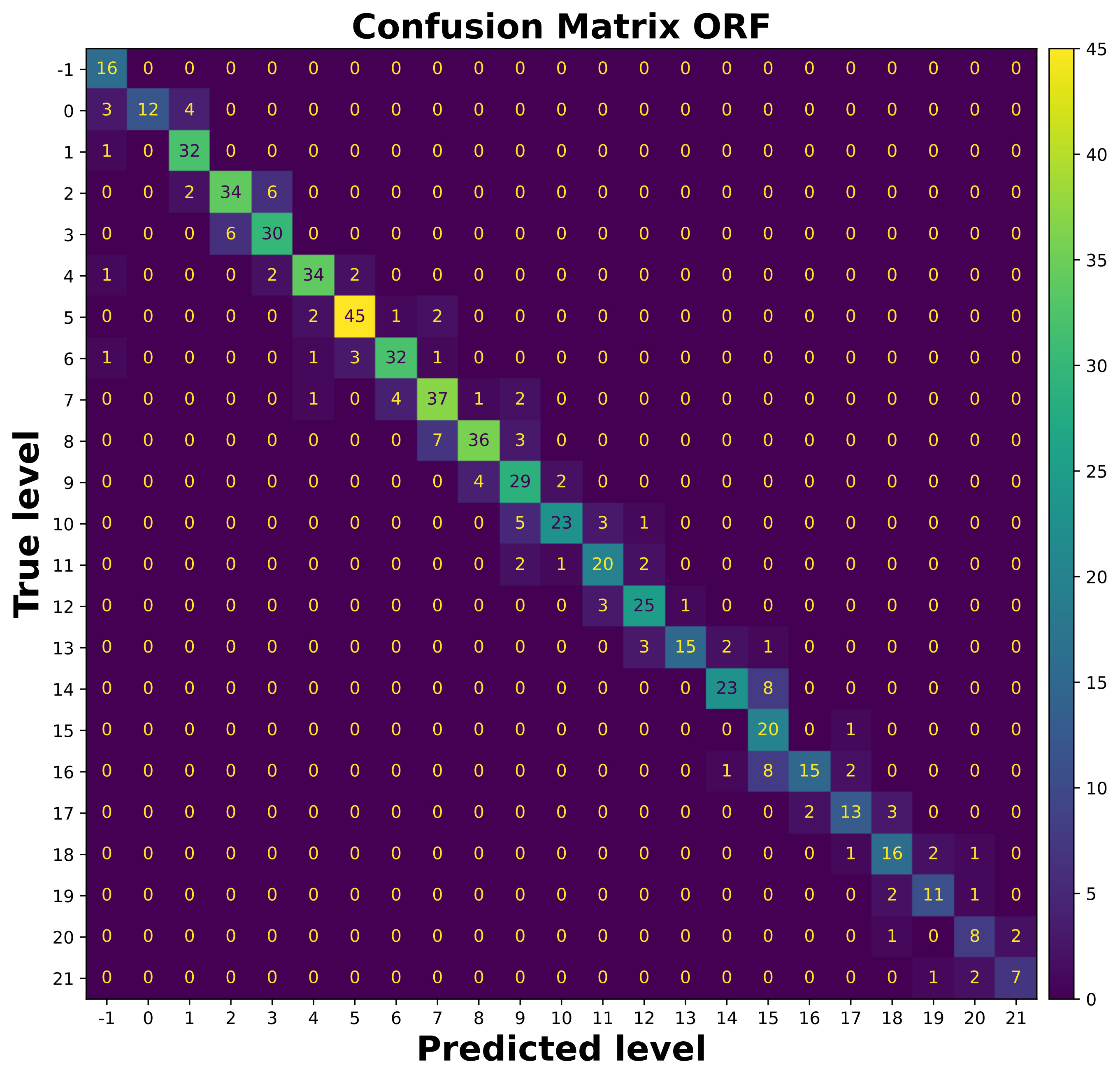}
    \caption{Confusion matrix - ORF}
    \label{fig:results-cm-orf}
\end{figure}

Interestingly, the ORF model is the only one that achieved a slightly better result for macro-averaged RMSE than for the regular RMSE. However, for MAE, its regular version still yields better performance. This can be attributed to the difficulty of the model in correctly predicting a few specific samples from the lower levels, particularly a monster for which the distance between the true and predicted label was as high as 7 levels, see Figure \ref{fig:results-cm-orf}. This single extreme misclassification substantially affected macroaveraged metrics. Although other models such as SAOC also misclassified this instance, their predictions were significantly closer to the true label.

Overall, macroaveraged metrics provide a more balanced and informative view of model performance across all levels, especially in the presence of class imbalance. Unlike regular metrics, they prevent overly optimistic evaluations driven by models that perform well only in the most frequent classes.

\subsection{Expanding window}

\subsubsection{Overall models performance comparison}
\label{sec:results_window_all}

\begin{table}[h]
    \centering
\begin{tabular}{|l||>{\centering\arraybackslash}m{2cm}|>{\centering\arraybackslash}m{2cm}|c|c|c|}
  \hline    
  \textbf{Model} & \textbf{$\mathbf{MAE}^{\mathbf{M}}$} & \textbf{$\mathbf{RMSE}^{\mathbf{M}}$} & \textbf{Somer's D} & \textbf{Acc} & \textbf{Acc@1} \\
  \hline
  \hline
      Baseline & \text{$0.99 \pm 0.42$} & \text{$1.56 \pm 0.55$} & \text{$0.87 \pm 0.07$} & \text{$39 \pm 10$}\% & \text{$78 \pm 12$}\% \\

      LR & \text{$0.37 \pm 0.12$} &\text{$0.70 \pm 0.15$} &\cellcolor{violet!25}\text{$0.95 \pm 0.02$} &\text{$66 \pm 8\%$} & \cellcolor{blue!25}\text{$97 \pm 2\%$} \\
      SVM &  \text{$0.35 \pm 0.15$} &\text{$0.68 \pm 0.21$} & \cellcolor{blue!25}\text{$0.96 \pm 0.02$} & \text{$71 \pm 1\%$} & \cellcolor{blue!25}\text{$97 \pm 3\%$} \\
      RF & \cellcolor{violet!25}\text{$0.26 \pm 0.12$} &\cellcolor{blue!25}\text{$0.56 \pm 0.19$} & \cellcolor{green!50}\text{$0.97 \pm 0.02$} &\cellcolor{blue!25}\text{$75 \pm 1\%$} & \cellcolor{green!50}\text{$98 \pm 2\%$} \\
      LightGBM & \text{$0.28 \pm 0.11$} & \cellcolor{violet!25}\text{$0.59 \pm 0.18$} & \cellcolor{green!50}\text{$0.97 \pm 0.02$} &\cellcolor{violet!25}\text{$73 \pm 10\%$} & \cellcolor{green!50}\text{$98 \pm 3\%$} \\
      \hline
      ORD & \text{$0.32 \pm 0.12$} &\text{$0.66 \pm 0.17$} & \cellcolor{blue!25}\text{$0.96 \pm 0.02$} &\text{$69 \pm 8\%$} & \cellcolor{blue!25}\text{$97 \pm 2\%$} \\
      ORF & \cellcolor{blue!25}\text{$0.25 \pm 0.10$} &\cellcolor{blue!25}\text{$0.56 \pm 0.14$} & \cellcolor{green!50}\text{$0.97 \pm 0.02$} & \cellcolor{blue!25}\text{$75 \pm 10\%$} & \cellcolor{blue!25}\text{$97 \pm 2\%$} \\
      LogisticIT & \text{$0.32 \pm 0.11$} &\text{$0.64 \pm 0.15$} & \cellcolor{blue!25}\text{$0.96 \pm 0.01$} &\text{$69 \pm 8\%$} & \cellcolor{blue!25}\text{$97 \pm 2\%$} \\
      SAOC & \cellcolor{green!50}\text{$0.24 \pm 0.12$} & \cellcolor{green!50}\text{$0.52 \pm 0.13$} & \cellcolor{green!50}\text{$0.97 \pm 0.02$} & \cellcolor{green!50}\text{$76 \pm 11\%$} & \cellcolor{blue!25}\text{$97 \pm 2\%$} \\
      GPOR & \text{$0.32 \pm 0.12$} & \text{$0.63 \pm 0.15$} & \cellcolor{blue!25}\text{$0.96 \pm 0.02$} & \text{$71 \pm 8$}\% & \text{$97 \pm 3$}\% \\
      CORAL & \text{$0.49 \pm 0.25$} &\text{$0.81 \pm 0.31$} &\cellcolor{violet!25}\text{$0.95 \pm 0.02$} &\text{$58 \pm 12\%$} &\cellcolor{violet!25}\text{$95 \pm 6\%$} \\
      CORN & \text{$0.48 \pm 0.19$} &\text{$0.82 \pm 0.25$} &\cellcolor{violet!25}\text{$0.95 \pm 0.02$} &\text{$58 \pm 11\%$} &\text{$94 \pm 5\%$} \\
      Spacecutter & \text{$0.52 \pm 0.16$} &\text{$0.84 \pm 0.18$} &\text{$0.93 \pm 0.03$} &\text{$52 \pm 7\%$} &\text{$94 \pm 5\%$} \\
      NNRank & \text{$0.70 \pm 0.83$} &\text{$1.08 \pm 1.04$} & \cellcolor{violet!25}\text{$0.95 \pm 0.02$} &\text{$56 \pm 16\%$} &\text{$88 \pm 20\%$} \\
      CONDOR & \text{$0.49 \pm 0.16$} &\text{$0.82 \pm 0.20$} &\cellcolor{violet!25}\text{$0.95 \pm 0.03$} &\text{$56 \pm 13\%$} &\text{$93 \pm 6\%$} \\
      OR-CNN & \text{$0.52 \pm 0.22$} &\text{$0.86 \pm 0.22$} &\cellcolor{violet!25}\text{$0.95 \pm 0.02$} &\text{$56 \pm 13\%$} &\text{$92 \pm 6\%$} \\
      \hline
\end{tabular}
    \caption{Models results chosen based on MAE macroaveraged.}
    \label{tab:results_window_models_all}
\end{table}

\begin{table}[h]
    \centering
\begin{tabular}{|l||>{\centering\arraybackslash}m{2cm}|>{\centering\arraybackslash}m{2cm}|c|c|c|}
  \hline    
  \textbf{Model} & \textbf{$\mathbf{MAE}^{\mathbf{M}}$} & \textbf{$\mathbf{RMSE}^{\mathbf{M}}$} & \textbf{Somer's D} & \textbf{Acc} & \textbf{Acc@1} \\
  \hline
  \hline
      SAOC & \text{$0.24 \pm 0.12$} &\text{$0.52 \pm 0.13$} &\text{$0.97 \pm 0.02$} &\text{$76 \pm 11\%$} &\text{$97 \pm 2\%$} \\
      ORF & \text{$0.25 \pm 0.10$} &\text{$0.56 \pm 0.14$} &\text{$0.97 \pm 0.02$} &\text{$75 \pm 10\%$} &\text{$97 \pm 2\%$} \\
      RF & \text{$0.26 \pm 0.12$} &\text{$0.56 \pm 0.19$} &\text{$0.97 \pm 0.02$} &\text{$75 \pm 1\%$} &\text{$98 \pm 2\%$} \\
      LightGBM & \text{$0.28 \pm 0.11$} &\text{$0.59 \pm 0.18$} &\text{$0.97 \pm 0.02$} &\text{$73 \pm 10\%$} &\text{$98 \pm 3\%$} \\
      GPOR & \text{$0.32 \pm 0.12$} &\text{$0.63 \pm 0.15$} &\text{$0.96 \pm 0.02$} &\text{$71 \pm 8\%$} &\text{$97 \pm 3\%$} \\
      \hline
\end{tabular}
    \caption{Top-5 models results based on MAE macroaveraged}
    \label{tab:results_window_models_top}
\end{table}

The general results presented in Table~\ref{tab:results_window_models_all} are largely consistent with previous evaluations. As expected, the baseline model performed significantly worse than all others in all metrics. In contrast, the best-performing models were consistently tree-based approaches - RF, LightGBM, ORF, and SAOC - demonstrating strong generalization and robustness, even when evaluated using an expanding window strategy.

A common challenge observed in most models is the slight degradation of performance, especially in the earlier windows, where the amount of training data was relatively limited. Despite this, tree-based models exhibited high resilience, achieving results that closely match those obtained under standard chronological train-test splitting. The most noticeable performance drop for these models was in raw accuracy, though even here they maintained top-tier rankings. Notably, the Somer's D metric remained both stable and consistently high across all models, indicating reliable ordinal consistency. Interestingly, for a subset of models, the windowed evaluation approach produced better results on average than the traditional train-test split. It suggests that these models may generalize more effectively than was shown for the previous setup.

Similarly to the results from the train-test split, there is no definitive winner between dedicated ordinal methods and regression models with rounding. However, the subgroup of neural network-based approaches continues to perform noticeably worse than other ordinal models.

One particularly interesting finding is the poor performance of the NNRank model. In one of the test windows (specifically, the eleventh), the model predicted a constant label for all instances, which resulted in extremely poor metrics, for example, macro-averaged RMSE exceeding 4 levels. This highlights the instability of some neural network models, especially those without mechanisms to enforce rank consistency or other improvement above a simple tactic. However, incorporating loss weighting into neural architectures (OR-CNN) showed some improvements. More sophisticated methods like CONDOR and CORN performed slightly better than OR-CNN and demonstrated more stable behavior (i.e. lower standard deviations across windows), though still underperformed compared to simpler classical methods.

\subsubsection{Accuracy}
\label{sec:results_window_acc}

\begin{table}[h]
    \centering
\begin{tabular}{|l||>{\centering\arraybackslash}m{2cm}|>{\centering\arraybackslash}m{2cm}|c|c|c|}
  \hline    
  \textbf{Model} & \textbf{Acc} & \textbf{Acc@1} \\
  \hline
  \hline
      Baseline  &\text{$39 \pm 10$}\% &\text{$78 \pm 12$}\% \\
      LR &\text{$66 \pm 8$}\% &\text{$97 \pm 2$}\% \\
      SVM &\text{$71 \pm 1$}\% &\text{$97 \pm 3$}\% \\
      RF &\text{$75 \pm 1$}\% &\text{$98 \pm 2$}\% \\
      LightGBM &\text{$73 \pm 10$}\% &\text{$98 \pm 3$}\% \\
      \hline
      ORD &\text{$69 \pm 8$}\% &\text{$97 \pm 2$}\% \\
      ORF &\text{$75 \pm 10$}\% &\text{$97 \pm 2$}\% \\
      LogisticIT &\text{$69 \pm 8$}\% &\text{$97 \pm 2$}\% \\
      SAOC &\text{$76 \pm 11$}\% &\text{$97 \pm 2$}\% \\
      GPOR &\text{$71 \pm 8$}\% &\text{$97 \pm 3$}\% \\
      CORAL &\text{$58 \pm 12$}\% &\text{$95 \pm 6$}\% \\
      CORN &\text{$58 \pm 11$}\% &\text{$94 \pm 5$}\% \\
      Spacecutter &\text{$52 \pm 7$}\% &\text{$94 \pm 5$}\% \\
      NNRank &\text{$56 \pm 16$}\% &\text{$88 \pm 20$}\% \\
      CONDOR & \text{$56 \pm 13$}\% &\text{$93 \pm 6$}\% \\
      OR-CNN & \text{$56 \pm 13$}\% &\text{$92 \pm 6$}\% \\
      \hline
\end{tabular}
    \caption{Models accuracy}
    \label{tab:results_window_acc}
\end{table}

As shown in Table~\ref{tab:results_window_acc}, for most models, accuracy@1 is noticeably more stable compared to standard accuracy. For the best-performing models, the standard deviation of accuracy@1 remains around 2–3 percentage points, whereas the standard accuracy often fluctuates by 10 or more points. This variation is slightly smaller for linear, logistic, and probabilistic models, i.e. LR, ORD, LogisticIT, and GPOR.

Although the mean values of standard accuracy are generally lower compared to the chronological train-test split, accuracy@1 remains relatively high and close to previous results, particularly for the top-performing models. The differences in accuracy@1 between models are somewhat larger than before, yet the majority still handle predictions well, with only some instances misclassified by more than one level. The key distinction between models often lies in their ability to achieve exact classifications and minimize large prediction errors.

\subsubsection{Rounding strategies comparison}

\begin{table}[h]
  \centering
  \resizebox{\textwidth}{!}{%
  \begin{tabular}{|l||c|c|c|c|c|c|c|}
    \hline
    \textbf{Model} &
    \makecell{\textbf{Round}\\\textbf{0.5}} &
    \makecell{\textbf{Global}\\\textbf{$R_1$}} &
    \makecell{\textbf{TPE}\\\textbf{$R_1$}} &
    \makecell{\textbf{Graph}\\\textbf{$R_1$}} &
    \makecell{\textbf{Global}\\\textbf{$R_2$}} &
    \makecell{\textbf{TPE}\\\textbf{$R_2$}} &
    \makecell{\textbf{Graph}\\\textbf{$R_2$}} \\
    \hline
    \hline
    Baseline & \text{$0.99 \pm 0.42$} & \text{$0.99 \pm 0.42$} & \text{$0.97 \pm 0.41$} & \cellcolor{green!50}\text{$0.94 \pm 0.39$} & \text{$0.99 \pm 0.42$} & \text{$0.98 \pm 0.41$} & \cellcolor{green!50}\text{$0.94 \pm 0.39$} \\
    LR & \text{$0.38 \pm 0.11$} & \text{$0.38 \pm 0.11$} & \text{$0.39 \pm 0.14$} & \text{$0.37 \pm 0.12$} & \text{$0.38 \pm 0.11$} & \cellcolor{green!50}\text{$0.37 \pm 0.12$} & \text{$0.38 \pm 0.12$} \\
    SVM & \text{$0.37 \pm 0.18$} & \text{$0.37 \pm 0.17$} & \text{$0.37 \pm 0.17$} & \cellcolor{green!50}\text{$0.35 \pm 0.15$} & \text{$0.37 \pm 0.17$} & \text{$0.38 \pm 0.17$} & \text{$0.35 \pm 0.16$} \\
    RF & \text{$0.29 \pm 0.13$} & \text{$0.28 \pm 0.11$} & \text{$0.30 \pm 0.12$} & \text{$0.26 \pm 0.11$} & \text{$0.28 \pm 0.11$} & \text{$0.29 \pm 0.13$} & \cellcolor{green!50}\text{$0.26 \pm 0.12$} \\
    LightGBM & \cellcolor{green!50}\text{$0.28 \pm 0.11$} & \text{$0.48 \pm 0.14$} & \text{$0.35 \pm 0.09$} & \text{$0.48 \pm 0.13$} & \text{$0.33 \pm 0.12$} & \text{$0.29 \pm 0.1$} & \text{$0.32 \pm 0.12$} \\
    \hline
  \end{tabular}
  }
  \caption{MAE rounding results comparison}
  \label{tab:results_window_rounding}
\end{table}

As shown in Table~\ref{tab:results_window_rounding}, the differences between the rounding strategies are generally small. Among the strategies evaluated, the graph-based rounding threshold method often achieves the best performance. However, LightGBM again suffers from overfitting, which negatively affects its performance with this more flexible thresholding. This model tends to use the lowest allowed threshold, indicating that setting a narrower threshold range may help mitigate its overfitting repercussions. As discussed in Section~\ref{sec:results_rounding}, this suggests that the current conflict resolution approach, choosing the first among equally optimal thresholds, is not optimal for overfitted models and may require revision.

Interestingly, for the LR model, the TPE-based rounding strategy yielded the best results. There is also no universal pattern among the results that will suggest that only one strategy could always be applied in all models. This reinforces the importance of tailoring rounding strategies to model-specific characteristics, as no single approach consistently dominates across all models.

It is also worth noting that the standard deviations across strategies remain relatively stable, suggesting consistent behavior during validation despite the different rounding configurations.

\subsubsection{Comparison of Regular and Macroaveraged}
\label{sec:regular_vs_macroaveraged_window}

\begin{table}[H]
    \centering
\begin{tabular}{|l||>{\centering\arraybackslash}m{2cm}|>{\centering\arraybackslash}m{2cm}|c|c|c|}
  \hline
  \textbf{Model} & \textbf{MAE} & \textbf{$\mathbf{MAE}^{\mathbf{M}}$} & \textbf{RMSE} & \textbf{$\mathbf{RMSE}^{\mathbf{M}}$} \\
  \hline
  \hline
    Baseline & \text{$1.03 \pm 0.45$} & \text{$0.99 \pm 0.42$} & \text{$1.59 \pm 0.58$} & \text{$1.56 \pm 0.55$} \\
    
    LR & \text{$0.39 \pm 0.09$} & \text{$0.37 \pm 0.12$} & \text{$0.73 \pm 0.14$} & \text{$0.70 \pm 0.15$} \\
    
    SVM & \text{$0.35 \pm 0.15$} & \text{$0.35 \pm 0.15$} & \text{$0.68 \pm 0.24$} & \text{$0.68 \pm 0.21$} \\
    
    RF & \text{$0.28 \pm 0.13$} & \text{$0.26 \pm 0.12$} & \text{$0.58 \pm 0.19$} & \text{$0.56 \pm 0.19$} \\
    
    LightGBM & \text{$0.29 \pm 0.13$} & \text{$0.28 \pm 0.11$} & \text{$0.60 \pm 0.19$} & \text{$0.59 \pm 0.18$} \\
    
    \hline
    
    ORD & \text{$0.34 \pm 0.09$} & \text{$0.32 \pm 0.12$} & \text{$0.68 \pm 0.14$} & \text{$0.66 \pm 0.17$} \\
    
    ORF & \text{$0.27 \pm 0.13$} & \text{$0.25 \pm 0.10$} & \text{$0.58 \pm 0.16$} & \text{$0.56 \pm 0.14$} \\
    
    LogisticIT & \text{$0.33 \pm 0.09$} & \text{$0.32 \pm 0.11$} & \text{$0.67 \pm 0.14$} & \text{$0.64 \pm 0.15$} \\
    
    SAOC & \text{$0.25 \pm 0.13$} & \text{$0.24 \pm 0.12$} & \text{$0.54 \pm 0.14$} & \text{$0.52 \pm 0.13$} \\
    
    GPOR & \text{$0.33 \pm 0.10$} & \text{$0.32 \pm 0.12$} & \text{$0.65 \pm 0.15$} & \text{$0.63 \pm 0.15$} \\
    
    CORAL & \text{$0.49 \pm 0.22$} & \text{$0.49 \pm 0.25$} & \text{$0.82 \pm 0.27$} & \text{$0.81 \pm 0.31$} \\
    
    CORN & \text{$0.49 \pm 0.20$} & \text{$0.48 \pm 0.19$} & \text{$0.83 \pm 0.26$} & \text{$0.82 \pm 0.25$} \\
    
    Spacecutter & \text{$0.56 \pm 0.13$} & \text{$0.52 \pm 0.16$} & \text{$0.87 \pm 0.15$} & \text{$0.84 \pm 0.18$} \\
    
    NNRank & \text{$0.73 \pm 0.90$} & \text{$0.70 \pm 0.83$} & \text{$1.09 \pm 1.02$} & \text{$1.08 \pm 1.04$} \\
    
    CONDOR & \text{$0.52 \pm 0.17$} & \text{$0.49 \pm 0.16$} & \text{$0.84 \pm 0.19$} & \text{$0.82 \pm 0.2$} \\
    
    OR-CNN & \text{$0.53 \pm 0.19$} & \text{$0.52 \pm 0.22$} & \text{$0.87 \pm 0.18$} & \text{$0.86 \pm 0.22$} \\
    
    \hline
\end{tabular}
    \caption{Regular and macroaveraged MAE and RMSE values.}
    \label{tab:results_window_rmse_mae}
\end{table}

\begin{table}[]
    \centering
    \begin{tabular}{|>{\centering\arraybackslash}p{2.6cm}||c|c|c|c|}
        \hline
        \textbf{Test window number} & \textbf{MAE} & \textbf{$\mathbf{MAE}^{\mathbf{M}}$} & \textbf{RMSE} & \textbf{$\mathbf{RMSE}^{\mathbf{M}}$} \\
        
        \hline
        \hline
        
        1 & 0.49 & 0.46 & 0.78 & 0.74 \\

        2 & 0.28 & 0.15 & 0.58 & 0.42 \\
        
        3 & 0.27 & 0.31 & 0.56 & 0.58 \\
        
        4 & 0.24 & 0.30 & 0.53 & 0.59 \\
        
        5 & 0.20 & 0.16 & 0.51 & 0.44 \\
        
        6 & 0.26 & 0.28 & 0.54 & 0.56 \\
        
        7 & 0.57 & 0.52 & 0.85 & 0.82 \\
        
        8 & 0.12 & 0.13 & 0.37 & 0.38 \\
        
        9 & 0.13 & 0.17 & 0.40 & 0.45 \\
        
        10 & 0.22 & 0.25 & 0.51 & 0.54 \\
        
        11 & 0.14 & 0.15 & 0.42 & 0.45 \\
        
        12 & 0.23 & 0.23 & 0.51 & 0.51 \\
        
        13 & 0.28 & 0.25 & 0.62 & 0.57 \\
        \hline
    \end{tabular}
    \caption{Regular and macroaveraged MAE and RMSE values of sliding test windows for SAOC}
    \label{tab:results_saoc_mae_rmse}
\end{table}

Interestingly, in this evaluation, no model exhibits higher error values for macroaveraged metrics compared to the regular ones, as shown in Table~\ref{tab:results_window_rmse_mae}. This outcome can be attributed to specific test windows, as seen in the SAOC results in Table~\ref{tab:results_saoc_mae_rmse}. In particular, this phenomenon is not isolated; it occurs in five separate windows across the dataset. These anomalies do not follow an obvious pattern: they do not coincide with local minima or maxima, nor are they present only for early training windows. Importantly, this behavior is not directly linked to the number of training samples, since it appears also for later windows. 

In some early windows, for example, certain classes are heavily underrepresented in the training set but present in larger numbers in the test set. For example, in the first window, there is only one monster at level 21 in training, but six such monsters appear in the test set. Additionally, some classes that are moderately represented in the overall dataset are underrepresented in the early sourcebooks, behaved similarly to rare classes in those windows, e.g. level -1 in the second window. Moreover, difficult-to-predict examples, an example discussed in Section~\ref{sec:regular_vs_macroaveraged}, end up in smaller test windows, increasing their impact on the metrics. 

These results demonstrate that the sliding window evaluation better reflects real-world deployment scenarios, where the distribution of new data is unlikely to match that of the training set. On the other hand, one might argue that severely imbalanced or atypical test sets may not provide a reliable assessment of a model's generalization ability, as rare classes can dominate certain windows.

Furthermore, fluctuations in both regular and macroaveraged metric values appear to follow similar trends across windows. This suggests that while both versions reflect the same overall performance pattern, including local minima and maxima, they do not necessarily vary by the same magnitude in each window, e.g. between window 2 and 3. In some cases, the regular metric yields better performance; in others, the macroaveraged metric performs better.

However, the findings strongly support the use of macroaveraged metrics to gain a deeper understanding of the performance of the model. At the same time, inspecting the standard metrics (regular MAE and RMSE) and the window results remains valuable for interpreting outcomes and better understanding both model behavior and dataset structure.

\subsubsection{Window results analysis}

\begin{table}[h]
    \centering
\begin{tabular}{|>{\centering\arraybackslash}m{2.6cm}||>{\centering\arraybackslash}m{2cm}|>{\centering\arraybackslash}m{2cm}|c|c|c|}
  \hline    
  \textbf{Test window number} & \textbf{$\mathbf{MAE}^{\mathbf{M}}$} & \textbf{$\mathbf{RMSE}^{\mathbf{M}}$} & \textbf{Somer's D} & \textbf{Acc} & \textbf{Acc@1} \\
  \hline
  \hline
    1 & 0.55 & 1.04 & 0.94 & 59\% & 91\% \\
    
    2 & 0.22 & 0.50 & 0.93 & 57\% & 94\%\\
    
    3 & 0.28 & 0.53 & 0.97 & 75\% & 100\%\\
    
    4 & 0.35 & 0.63 & 0.97 & 69\% & 98\%\\
    
    5 & 0.10 & 0.31 & 0.97 & 85\% & 100\%\\
    
    6 & 0.34 & 0.81 & 0.97 & 77\% & 97\% \\
    
    7 & 0.33 & 0.66 & 0.95 & 63\% & 99\% \\
    
    8 & 0.19 & 0.46 & 0.99 & 85\% & 99\% \\
    
    9 & 0.15 & 0.39 & 0.99 & 84\% & 100\% \\
    
    10 & 0.24 & 0.53 & 0.98 & 76\% & 98\% \\
    
    11 & 0.22 & 0.53 & 0.98 & 81\% & 98\% \\
    
    12 & 0.29 & 0.62 & 0.97 & 75\% & 97\% \\
    
    13 & 0.10 & 0.34 & 0.98 & 88\% & 99\% \\
    \hline
\end{tabular}
    \caption{Detailed results of sliding test windows for RF model (graph rounding with $R_2$ thresholds).}
    \label{tab:results_window_example_rf}
\end{table}

There is a general tendency for models to perform better when a larger number of monsters are available in the training data, see example results for RF with graph rounding $R_2$ thresholds in Table \ref{tab:results_window_example_rf}. However, this relationship is not strictly linear and more data does not always guarantee better results. The worst performance was observed in the first test window, where the training set was particularly limited. In some cases, certain levels (e.g., level 21) were nearly absent, which significantly hindered model performance. Local improvements in performance metrics were observed in some windows; there were also instances of local degradation. It is important to note that the monster data are not independent and identically distributed (i.i.d.), partly due to temporal dependencies. This non-i.i.d. nature may contribute to the observed local performance fluctuations across windows.

It is important to note that, due to the temporal nature of the data, the distribution of levels varies across different time windows, as described in Section \ref{sec:regular_vs_macroaveraged_window}. Not all levels are represented in each test set. Because the data is based on a chronological publishing process, differences between books and between sets of books are influenced by the activity of the publisher. The time gaps between publications, as well as the design decisions made by specific creators, introduce additional variability. A change in the designer within the same company can noticeably affect the data, as individual design styles influence the characteristics of the monster. Given all these factors that are human-driven and time dependent, it is notable that the models manage to maintain strong performance, even in the early stages of training.
\chapter{Summary}

This thesis presented a comprehensive study on the application of ML techniques for estimating monster levels in the context of pen \& paper RPG game design, specifically for Pathfinder Roleplaying Game: Second Edition. The goal of this work was to develop an accurate and efficient automated system capable of predicting monster levels, thus supporting game designers and Game Masters in their workflow.

To enable this, a dedicated dataset was constructed using information extracted from the official Pathfinder rulebooks. The dataset includes a wide range of features that represent the attributes, abilities, and statistics of the monsters. It was carefully engineered to reflect the inherent chronological order of the source material.

Various ML models suitable for the ordinal regression problem were explored and implemented. These included both classic regression models adapted with rounding strategies and specialized models explicitly designed to handle ordinal data. Particular attention was paid to ensure that the evaluation process reflected the real-world scenario of deploying models on newly published content. As such, the evaluation framework was built around chronological data splits and expanding window validation, designed to prevent data leakage and better measure generalization performance over time.

The experiments demonstrated that, when appropriately tuned and evaluated, machine learning models can provide reliable estimations of the level of the monster. This opens up potential for practical applications in RPG content development, such as balancing encounters and generating new monsters.

\section{Future Work}

There are several promising directions for further development that build on this work. One key improvement would be to use the validation set specifically to adjust the rounding threshold. Currently, thresholds are optimized using predictions in the training set, which can be problematic for models prone to overfitting. In such cases, the predicted values on the training data tend to focus very closely around the true labels, resulting in thresholds that are biased toward integer or class values. Using a validation set would help mitigate this effect and lead to more generalizable threshold selections.

Additional experimentation with alternative loss functions for neural networks tailored specifically to ordinal regression tasks could also be explored. The existing work briefly investigated such approaches, but further refinements may yield improved results.

From a usability standpoint, the integration of explainable AI (XAI) methods, such as counterfactual explanations, could provide significant value. These techniques could assist game designers in understanding which monster attributes most influence level prediction. They can also guide them in adjusting monster characteristics when a specific level target is desired, while preserving the intended narrative or gameplay role of the creature.

\section{Potential Applications}

The market for pen \& paper PRG games is growing rapidly, and the broader trend toward AI-assisted design tools is making its way into this domain. The models developed in this thesis have direct practical applications that can improve performance and enhance the game design process.

One key use case is to shorten the development cycle for new monsters. Currently, balancing the level of a monster requires extensive playtesting, often spanning multiple sessions of 2 to 6 hours each. Using the trained models from this work as an initial estimator, designers can potentially bypass or reduce the number of these sessions, accelerating the development process while maintaining quality.

Another application is for Game Masters (GMs) who create custom monsters for player-driven adventures. Unlike professional designers, GMs often lack the time and resources for a comprehensive playtest. As demonstrated in this thesis, even guideline-based intuitive methods provided by the official Pathfinder materials may yield insufficient results. Integrating the trained model into a user-friendly application could significantly support GMs in designing balanced encounters, especially for custom adventures.

A prototype application has already been implemented for personal use and has proven effective in improving session balance and overall player experience. Expanding this into a public tool, possibly integrated with other game design features, could provide significant value to both hobbyists and professional content creators in the RPG community.


\printbibliography  
\end{document}